\documentclass[times, review, 10pt]{elsarticle}

\usepackage{amsfonts}
\usepackage{amsmath,amssymb,amsthm}
\usepackage{bbold}
\usepackage{mathtools}
\usepackage{algorithm,algorithmic}
\usepackage{microtype}

\usepackage[caption=false,font=footnotesize]{subfig}
\usepackage{graphicx}				
\usepackage{booktabs}				
\usepackage{array}					
\usepackage{multirow}				
\usepackage{siunitx}				
\usepackage{flafter}				
\usepackage{comment}				
\usepackage[dvipsnames]{xcolor}	
\usepackage{flafter}				
\usepackage{lipsum} 				
\usepackage{setspace}               
\usepackage{enumitem}
\usepackage{multicol}

\usepackage{amssymb}
\usepackage{soul}
\usepackage{url}
\usepackage{epstopdf}
\usepackage{rotating}
\usepackage{placeins}
\usepackage{pifont}
\usepackage{threeparttable}
\usepackage{makecell}
\usepackage{float}
\usepackage{siunitx}

\usepackage[english]{babel}
\usepackage{hyperref}	
\usepackage{cleveref}	

\newtheorem{theorem}{Theorem}[section]

\usepackage{amsmath, colortbl}

\allowdisplaybreaks

\journal{Pattern Recognition}

\begin{document}

\begin{frontmatter}



\title{Wasserstein-based Kernel Principal Component Analysis for Clustering Applications}


\author[RRE]{Alfredo Oneto}

\author[RRE]{Blazhe Gjorgiev}

\author[RRE]{Giovanni Sansavini\corref{cor1}}
\ead{sansavig@ethz.ch}

\affiliation[RRE]{organization={Reliability and Risk Engineering Lab, Institute of Energy and Process Engineering, Department of Mechanical and Process Engineering, ETH Z\"urich},
            country={Switzerland}}

\cortext[cor1]{Corresponding author.}

\begin{abstract}
Many data clustering applications must handle objects that cannot be represented as vectors. In this context, the bag-of-vectors representation describes complex objects through discrete distributions, for which the Wasserstein distance provides a well-conditioned dissimilarity measure. Kernel methods extend this by embedding distance information into feature spaces that facilitate analysis. However, an unsupervised framework that combines kernels with Wasserstein distances for clustering distributional data is still lacking. We address this gap by introducing a computationally tractable framework that integrates Wasserstein metrics with kernel methods for clustering. The framework can accommodate both vectorial and distributional data, enabling applications in various domains. It comprises three components: (i) an efficient approximation of pairwise Wasserstein distances using multiple reference distributions; (ii) shifted positive definite kernel functions based on Wasserstein distances, combined with kernel principal component analysis for feature mapping; and (iii) scalable, distance-agnostic validity indices for clustering evaluation and kernel parameter optimization. Experiments on power distribution graphs and real-world time series demonstrate the effectiveness and efficiency of the proposed framework.
\end{abstract}


\end{frontmatter}


\section{Introduction} \label{sec:intro}


Data clustering is a fundamental problem in data science and has therefore attracted considerable attention. Data clustering aims to group similar objects while separating dissimilar ones into distinct clusters~\cite{WANG2024111062}. However, the intricacies associated with clustering pose relevant challenges. One challenge is defining the permissible cluster shapes, especially in datasets with complex and diverse structures. Additionally, clustering requires a suitable notion of similarity that aligns with the dataset’s characteristics. As data clustering advances to complex objects, the quest for solutions to these challenges continues, prompting ongoing research. 


A common approach to clustering involves learning a set of prototype objects to minimize the distances between them and the clustered objects. This approach is utilized by the popular \texttt{k-means}~\cite{kmeanslloyd1982} and \texttt{k-medoids}~\cite{kmedoidsrdusseeun1987}. \texttt{K-means} uses centroids as prototypes, whereas \texttt{k-medoids} uses medoids. These algorithms have been predominantly applied to vector spaces. Nevertheless, many tasks require the analysis of datasets with non-vectorial representations, for instance: electroencephalography signal processing~\cite{liu2025}, resting-state functional magnetic resonance imaging~\cite{zhu2022stacked, yuan2025}, and power Fourier spectra analysis~\cite{Cazelles2021}.

Adaptations of \texttt{k-means} and \texttt{k-medoids} have been developed for non-vectorial settings. In particular, extensions of these approaches to bag-of-vectors settings have attracted significant attention~\cite{okano2025wasserstein}, as they appear in several tasks involving objects such as images and graphs. In the bag-of-vectors representation, the information from individual elements of objects, such as nodes in graphs or pixels in images, is represented with weighted vectors. As these objects can be succinctly described through probability distributions, the use of Wasserstein distances and optimal transport techniques has provided rich geometrical tools for their clustering~\cite{hao2024clustering, delbarrio2019robust}.

Kernel methods incorporating Wasserstein distance information have extended the applicability of optimal transport methods in the machine learning domain~\cite{togninalli2019wasserstein, ma2023positive}. As kernels enable the representation of dot products in high-dimensional feature spaces, they facilitate data analysis and the combination of different similarity metrics~\cite{RAHIMZADEHARASHLOO2024110189}. Although past works have addressed problems of bag-of-vectors clustering and Wasserstein-based kernel learning, existing methods generally fall into two categories: clustering distributional information through non-kernel methods or kernel learning in supervised or semi-supervised settings. This gap motivates the need for an unsupervised framework for clustering that can treat distributional data through kernels. Such a framework enables the detection of patterns that would be unobservable in the data domain and the integration with other kernel measures over non-distributional information. This work offers the following main contributions:
\begin{enumerate}[label=(\roman*)]
    \item We introduce a multiple reference distribution approximation for pairwise Wasserstein distances, which enhances the existing single distribution approximation linear optimal transport method. This method reduces errors and diminishes the impact of reference selection, while maintaining computational tractability.
    \item We propose applying a constant positive shift to exponential kernels based on Wasserstein distances, which are indefinite, to obtain positive definite (p.d.) kernels. Moreover, we prove the invariance of clustering assignments under shifts. We use these kernels to derive feature maps through scalable kernel principal component analysis (PCA) methods that control the effects of low-variance components.
    \item We develop an efficient distance-agnostic validity index that compares the quality of cluster shapes. This index enables kernel parameters optimization by evaluating clustering configurations across different spaces, and we optimize them via Bayesian optimization.
\end{enumerate}

The remainder of this article is organized as follows. Section~\ref{sec:relatedworks} presents related work in essential aspects of our framework. Section~\ref{sec:method} develops the clustering framework. Section~\ref{sec:experiments} provides experimental results and analyses of clustering applied to time series and power distribution graphs. Section~\ref{sec:discussion} discusses the strengths and limitations of the proposed framework. Finally, Section~\ref{sec:conclusions} gives concluding remarks.

\section{Related works} \label{sec:relatedworks}

This section presents existing works on four relevant topics to our methodology.

\subsection{Computational methods for Wasserstein distances}

The Wasserstein distance is a metric for probability measures, which, despite its usefulness, is computationally expensive as it has no general closed-form solution. Only two exceptional cases have known closed-form solutions: the case of one-dimensional distributions~\cite{Cazelles2021} and the case of multi-dimensional location-scale families~\cite{verdinelli2019hybridwasserstein}, of which the Gaussian family is the most studied. In the literature, advancements have focused primarily on one of two types of complexity: speeding up the distance computation between distributions with a support of thousands of points or more, and reducing the complexity of calculating many pairwise distances in large datasets. For the former, the Sinkhorn and Sliced Wasserstein distances are the two approaches that can successfully treat very complicated distributions~\cite{peyre2019computational}. On the other hand, the linear optimal transport approach reduces the number of calculations to obtain pairwise Wasserstein distances in a dataset through a reference distribution~\cite{kolouri2021wasserstein}. This method has been  used for applications that require information on many pairwise relationships, such as clustering~\cite{verdinelli2019hybridwasserstein} and classification~\cite{kolouri2021wasserstein}.

\subsection{Clustering probability distributions}

In~\cite{puccetti2020computation}, the authors propose a \texttt{k-means} method based on Wasserstein distances, utilizing an iterative swapping algorithm. The work in~\cite{delbarrio2019robust} introduces a Wasserstein-based trimmed \texttt{k-means} that allows the trimming of the most discrepant distributions to improve computational performance and robustness. A fast distribution clustering method is introduced in~\cite{verdinelli2019hybridwasserstein}, which uses a hybrid measure approximating Wasserstein distances through the Gaussian closed-form solution. In~\cite{okano2025wasserstein}, the authors cluster probability distributions using principal geodesic analysis and the \texttt{k-centers} approach for functional data. Lastly,~\cite{hao2024clustering} proposes a distributional representation of chemical compounds to detect groups of compounds through their Wasserstein distances and the \texttt{DBSCAN} algorithm.

\subsection{Kernels incorporating Wasserstein distances}

Kernel methods are well-established in machine learning due to their computational tractability and their ability to reveal patterns that would otherwise be unobservable in the original data space~\cite{RAHIMZADEHARASHLOO2024110189}. These methods have also been applied to problems involving Wasserstein distances. For instance,~\cite{togninalli2019wasserstein} constructs an SVM classifier for graphs with a p.d. Wasserstein Laplacian kernel over categorical data. The work of~\cite{de2020wasserstein} proposes using Wasserstein exponential kernels for shape classification on small-scale datasets and approximates them to address indefiniteness. Moreover,~\cite{ma2023positive} constructs a graph kernel based on the graph sliced Wasserstein distance to classify brain functional networks.

\subsection{Validity indices for clustering evaluation}

Various validity indices have been proposed to compare and evaluate cluster configurations~\cite{WIROONSRI2024109910}, with a clear distinction between internal and external validation. The internal validity indices analyze only the clustered data. Most consist of functions that measure cluster cohesion and separation, such as the Davies-Bouldin and Silhouette indices. Some internal validity indices utilize counts of pointwise relations, such as the Goodman-Kruskal index (also known as the Gamma index) or the Tau index~\cite{TODESCHINI2024105117}. Others rely on information-theoretic measures, which allow for identifying robust clustering solutions that exhibit small diversity under different initializations~\cite{ni2023enhancing}. In kernel clustering, internal validity indices may fail because kernel parameters alter the intrinsic dimensionality of the feature space, which can substantially affect the outcome~\cite{tomavsev2016clustering}. External validity indices, such as accuracy and purity, evaluate partition quality against a known ground-truth partition, and kernel clustering applications commonly use them~\cite{SU2024kercorrelation}.


\section{Methodology}
\label{sec:method}

In this section, we present our clustering framework. Section~\ref{sec:wassdist} describes the Wasserstein distance and proposes a computationally tractable method for its approximation. Section~\ref{sec:featurespace} introduces the Wasserstein-based kernels, studies their properties, and presents the requirements for well-conditioned kernel principal component analysis. Section~\ref{sec:clustering} outlines a clustering method using the defined kernels and proposes suitable validity indices for optimizing kernel parameters.

\subsection{Wasserstein distances} \label{sec:wassdist}

The Wasserstein distance is a measure of the dissimilarity between two probability distributions. It quantifies the minimum transport cost needed to transform one probability distribution into the other. Consequently, it is suitable for comparing objects represented as distributions. The analytical definition of the Wasserstein distance with Euclidean transport costs between two probability distributions $\boldsymbol{\mu}_i$ and $\boldsymbol{\mu}_j$ defined on $\mathcal{X} \subseteq \mathbb{R}^d$, is~\cite{ma2023positive}:
\small
\begin{align*}
    W(\boldsymbol{\mu}_i, \boldsymbol{\mu}_j) = \left( \inf_{\boldsymbol{\pi} \in 
    \Pi (\boldsymbol{\mu}_i, \boldsymbol{\mu}_j)} \int_{\mathcal{X} \times \mathcal{X}} (||\boldsymbol{x}-\boldsymbol{y}||_2)^{2} d\pi(\boldsymbol{x},\boldsymbol{y}) \right)^{1/2} \tag{1} \label{eq1}
\end{align*}
\normalsize

\noindent where, $\mathcal{X}$ is a given normed vector space, $\Pi(\boldsymbol{\mu}_i, \boldsymbol{\mu}_j)$ denotes the set of all joint probability distributions, $\boldsymbol{\pi}$ is a joint distribution for $\boldsymbol{x}$ and $\boldsymbol{y}$ commonly referred to as a \textit{transport plan}, and $(\boldsymbol{x}, \boldsymbol{y})$ is any pair of points in $\mathcal{X} \times \mathcal{X}$. Given a pair $(\boldsymbol{x}, \boldsymbol{y})$, the value of $\boldsymbol{\pi}(\boldsymbol{x}, \boldsymbol{y})$ indicates the mass of $\boldsymbol{\mu}_i$ at $\boldsymbol{x}$ that is transported to $\boldsymbol{y}$ to reconstruct $\boldsymbol{\mu}_j$ from $\boldsymbol{\mu}_i$. The distance $W(\cdot,\cdot)$ in Eq.~(\ref{eq1}) is symmetric, nonnegative, satisfies the triangle inequality, and is finite when the second moments of both $\boldsymbol{\mu}_i$ and $\boldsymbol{\mu}_j$ exist.

Solving Eq.~(\ref{eq1}) is challenging, and numerical approaches are needed for the general case. Particularly, when $\boldsymbol{\mu}_i$ and $\boldsymbol{\mu}_j$ are discrete, the problem is reduced to solving an optimization model with a finite, possibly large, number of constraints and variables. We delve into the discrete settings for solving Eq.~(\ref{eq1}) in what follows.

\subsubsection{Discrete setting} \label{sec:wassdiscretesetting}

The Wasserstein distance between two discrete distributions $\boldsymbol{\mu}_i$ and $\boldsymbol{\mu}_j$ can be computed using optimal transport tools. Consider two bag-of-vectors in $\mathbb{R}^d$, $\boldsymbol{X} = [\boldsymbol{x}^m]_{m=1}^{N^{\boldsymbol{\mu}_i}}$ and $\boldsymbol{Y} = [\boldsymbol{y}^n]_{n=1}^{N^{\boldsymbol{\mu}_j}}$, where $N^{\boldsymbol{\mu}_i}$ and $N^{\boldsymbol{\mu}_j}$ represent the number of vectors. The vectors in $\boldsymbol{X}$ and $\boldsymbol{Y}$ have positive probabilities $p_m$ and $q_n$ such that $\sum_{m} p_m = 1$ and $\sum_{n} q_n = 1$.  We then define the discrete distributions supported on $\mathbb{R}^d$ as $\boldsymbol{\mu}_i = \sum_{m=1}^{N^{\boldsymbol{\mu}_i}} p_m \boldsymbol{\delta}_{\boldsymbol{x}^m}$ and $\boldsymbol{\mu}_j = \sum_{n=1}^{N^{\boldsymbol{\mu}_j}} q_n \boldsymbol{\delta}_{\boldsymbol{y}^n}$, where $\boldsymbol{\delta}_{\boldsymbol{x}^m}$ and $\boldsymbol{\delta}_{\boldsymbol{y}^n}$ are Dirac distributions at the locations $\boldsymbol{x}^m$ and $\boldsymbol{y}^n$, respectively. The Wasserstein distance between $\boldsymbol{\mu}_i$ and $\boldsymbol{\mu}_j$ can be obtained by solving the following linear program:
\small
\begin{align*}
    W(\boldsymbol{\mu}_i, \boldsymbol{\mu}_j)^2 &= \min_{\boldsymbol{\pi} \geq \boldsymbol{0}} \sum_{m=1}^{N^{\boldsymbol{\mu}_i}} \sum_{n=1}^{N^{\boldsymbol{\mu}_j}} \pi_{mn} \left( || \boldsymbol{x}^m - \boldsymbol{y}^n||_2 \right)^2 
 \tag{2a} \label{eq2a}\\
    \text{s.t.} \quad \sum_{m=1}^{N^{\boldsymbol{\mu}_i}} &\pi_{mn} = q_n  \quad \forall n = 1, \ldots, N^{\boldsymbol{\mu}_j} \quad \text{;} \quad \sum_{n=1}^{N^{\boldsymbol{\mu}_j}} \pi_{mn} = p_m  \quad \forall m = 1, \ldots, N^{\boldsymbol{\mu}_i}, \tag{2b} \label{eq2b}
\end{align*}
\normalsize

\noindent where $\pi_{mn}$ is the \textit{matching weight} between the support vectors $\boldsymbol{x}^m$ and $\boldsymbol{y}^n$. Hence, Eqs.~(\ref{eq2a})-(\ref{eq2b}) optimize the matching weights to minimize the sum of the transport costs to reconstruct $\boldsymbol{\mu}_j$ from $\boldsymbol{\mu}_i$.  
Although the problem is linear, its exact computation is challenging and scales at least as $\mathcal{O}(N^{3} \log(N))$ when comparing two distributions of $N$ support vectors. Furthermore, consider a set of discrete distributions $\Delta = \{\boldsymbol{\mu}_l\}_{l=1}^S$, where each distribution $\boldsymbol{\mu}_l$ represents $\sum_{m=1}^{N^{\boldsymbol{\mu}_l}} p_m^l \boldsymbol{\delta}_{\boldsymbol{x}^{m,l}}$, with $\boldsymbol{x}^{m,l}$ being the $m$-th support vector of the $l$-th distribution, and $p_m^l$ is its corresponding probability. Computing the $S(S-1)/2$ pairwise Wasserstein distances for these $S$ distributions in $\Delta$ requires solving $S(S-1)/2$ optimization problems and becomes rapidly intractable. This motivates the exploration of approximate solutions.


\subsubsection{Approximating pairwise Wasserstein distances} \label{sec:approxwass}

\begin{figure}[ht]
\centering
\captionsetup{justification=centering}
    \includegraphics[width=0.48\textwidth]{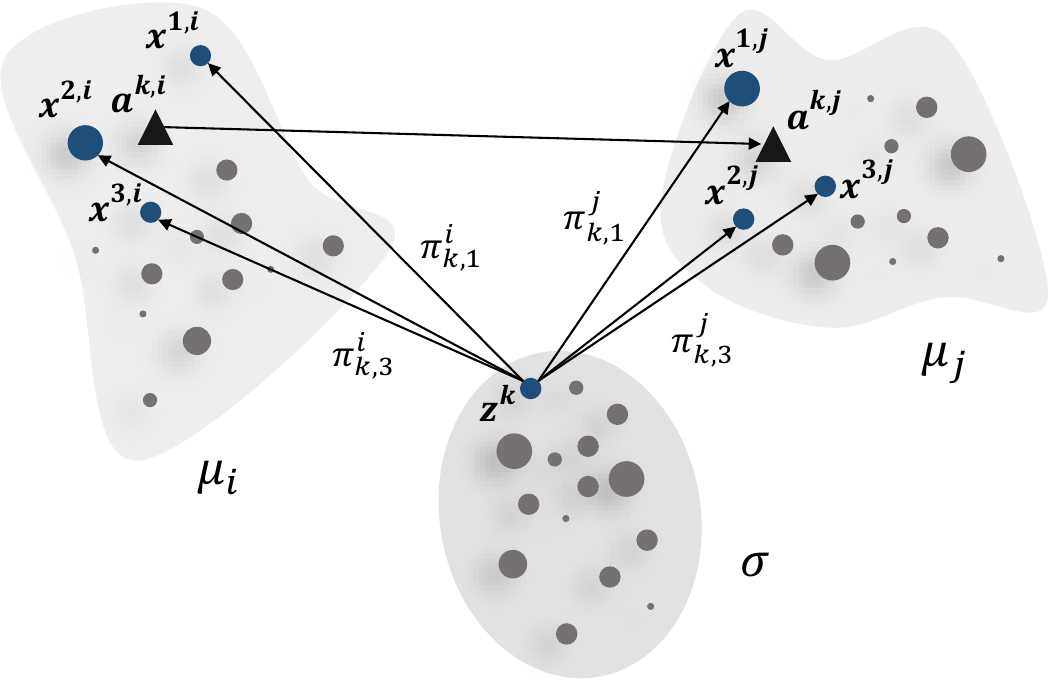}
    \caption{Illustration of the linear optimal transport distance variables. The reference $\boldsymbol{\sigma}$ is used to approximate the Wasserstein distance between the distributions $\boldsymbol{\mu}_i$ and $\boldsymbol{\mu}_j$.}
\label{fig:lotdistancev2}
\end{figure}

Several works employ the concepts of the linear optimal transportation framework due to its computational efficiency and practical success~\cite{kolouri2021wasserstein, verdinelli2019hybridwasserstein}. This framework approximates pairwise Wasserstein distances between distributions in a set $\Delta$ using a reference distribution $\boldsymbol{\sigma} = \sum_{k=1}^{N^{\boldsymbol{\sigma}}} g_k \boldsymbol{\delta}_{\boldsymbol{z}^k}$. Given the reference distribution $\boldsymbol{\sigma}$, we compute $W(\boldsymbol{\mu}_l, \boldsymbol{\sigma})$ with Eqs.~(\ref{eq2a})-(\ref{eq2b}) for every distribution in $\Delta$. The corresponding matching weights are used to compute the centroids of the mass transported from $\boldsymbol{z}^k$ to the distributions $\boldsymbol{\mu}_l$, respectively, which are referred to as \textit{forward images}:
\small
\begin{align*}
    \boldsymbol{a}^{k,l} = \frac{1}{g_k} \sum_{m=1}^{N^{\boldsymbol{\mu}_l}} \pi_{k, m}^l \boldsymbol{x}^{m,l} \qquad \forall k=1,\ldots,N^{\boldsymbol{\sigma}}, l=1,\ldots, S, \tag{3} \label{eq3}
\end{align*} 
\normalsize
which are then used to compute pairwise distances in $\Delta$:
\small
\begin{align*}
    L_{\boldsymbol{\sigma}}(\boldsymbol{\mu}_i, \boldsymbol{\mu}_j)^2 = \sum_{k=1}^{N^{\boldsymbol{\sigma}}} g_k   \left(||\boldsymbol{a}^{k,i} - \boldsymbol{a}^{k,j} ||_2 \right)^2 \qquad \forall (\boldsymbol{\mu}_i, \boldsymbol{\mu}_j) \in \Delta \times \Delta. \tag{4} \label{eq4}
\end{align*}
\normalsize

Fig.~\ref{fig:lotdistancev2} provides a graphical depiction of Eq.~(\ref{eq4}), which approximates $W(\cdot,\cdot)$ between any two distributions in the set $\Delta$\footnote{This approximation is not a proper metric, as in some cases $L_{\boldsymbol{\sigma}}(\boldsymbol{\mu}_i, \boldsymbol{\mu}_j) = 0$ for $\boldsymbol{\mu}_i \neq \boldsymbol{\mu}_j$.}. This distance denotes the transport cost required to move the forward image of each supporting vector $\boldsymbol{z}^k$ in the reference $\boldsymbol{\sigma}$, projected in $\boldsymbol{\mu}_i$, to the forward image in $\boldsymbol{\mu}_j$. This approximation allows us to obtain the pairwise distances of the dataset $\Delta$ with $S$ distributions by solving only $S$ optimization problems Eqs.~(\ref{eq2a})-(\ref{eq2b}), instead of $S(S-1)/2$. This is because the forward images $\boldsymbol{a}^{k,l}$ in Eq.~(\ref{eq3}) are obtained through the computation of $W(\boldsymbol{\mu}_l, \boldsymbol{\sigma})$ for $l=1, \ldots, S$. Then, the calculations described by Eq.~(\ref{eq4}) involve only weighted pairwise Euclidean distances.

\subsubsection{Approximation using multiple reference distributions} \label{sec:refselection}


A reference distribution far from the analyzed distributions reduces the quality of the distance approximation. In this context, we develop a method that leverages multiple $R$ reference distributions. The individual references are constructed using \texttt{k-means} clustering over the combined support vectors of the discrete distributions~\cite{kolouri2021wasserstein}. The multiple reference method employs the following steps:
\begin{itemize}
    \item \textit{Step 1 - Build an initial reference $\boldsymbol{\sigma}_1$}: We stack the support vectors of the distributions in $\Delta$ into a matrix $\boldsymbol{X}^{S}$. Subsequently, \texttt{k-means} is applied on the weighted $\boldsymbol{X}^{S}$ to determine an initial reference $\boldsymbol{\sigma}_1$. The number of centroids is set to $\left\lfloor \frac{1}{S} \sum_{l=1}^S N^{\boldsymbol{\mu}_l} \right\rfloor$, which is the lower approximation of the average number of support vectors in $\Delta$. Using $\boldsymbol{\sigma}_1$, we compute the $S(S-1)/2$ pairwise distances $\boldsymbol{D}_1 = \left[L_{\boldsymbol{\sigma}_1}(\boldsymbol{\mu}_i, \boldsymbol{\mu}_j)\right]_{i,j}$ in $\Delta$ via  Eq.~(\ref{eq4}). 
    
    \item \textit{Step 2 - Select additional references $\boldsymbol{\sigma}_r$}: We select $R-1$ additional distributions in $\Delta$. This is done by clustering the distributions in $\Delta$ into $R-1$ clusters with \texttt{k-medoids}, using the distances $\boldsymbol{D}_1$. The medoids are the additional reference distributions, $\boldsymbol{\sigma}_r$, for $r = 2, \ldots, R$.
\end{itemize}

Once the $R$ references have been selected, we compute all the pairwise distance matrices $\boldsymbol{D}_r$ corresponding to the distributions $\boldsymbol{\sigma}_r$ for $r = 2, \dots, R$.  Conveniently, the computation of the $\boldsymbol{D}_r$ matrices can be performed in parallel. Let $\Delta^{-}$ be the set defined as $\Delta \setminus \{\boldsymbol{\sigma}_2, \ldots, \boldsymbol{\sigma}_R\}$, i.e., all the distributions in $\Delta$ except the references selected in \textit{Step 2}. Then, the pairwise distance matrices are computed as:
\small
\begin{align*}
    D_{r,(\boldsymbol{\mu}_i, \boldsymbol{\mu}_j)} &= W(\boldsymbol{\mu}_i, \boldsymbol{\mu}_j) \qquad \forall (\boldsymbol{\mu}_i, \boldsymbol{\mu}_j) \in \Delta \times \Delta \text{ and }  \boldsymbol{\mu}_i = \boldsymbol{\sigma}_r \text{ or } \boldsymbol{\mu}_j = \boldsymbol{\sigma}_r \tag{5a} \label{eq5pa}\\
    D_{r,(\boldsymbol{\mu}_i, \boldsymbol{\mu}_j)} &= L_{\boldsymbol{\sigma}_r}(\boldsymbol{\mu}_i, \boldsymbol{\mu}_j) \qquad \forall (\boldsymbol{\mu}_i, \boldsymbol{\mu}_j) \in \Delta^{-} \times \Delta^{-}.\tag{5b} \label{eq5pb}
\end{align*}
\normalsize

Therefore, the distances between the reference $\boldsymbol{\sigma}_r$ and the other distributions in $\Delta$ are computed exactly (Eq.~(\ref{eq5pa})). Conversely, Eq.~(\ref{eq5pb}) approximates the distance between every two distributions $\boldsymbol{\mu}_i$ and $\boldsymbol{\mu}_j$ in $\Delta$ that do not belong to the reference distribution set $\boldsymbol{\sigma}_r$ using $L_{\boldsymbol{\sigma}_r}$. The pairwise distance matrix of $\Delta$, $\boldsymbol{\hat{D}}$, is obtained using the distance matrices $\boldsymbol{D}_1, \ldots, \boldsymbol{D}_R$:
\small
\begin{align*}
    \hat{D}_{\boldsymbol{\mu}_i, \boldsymbol{\mu}_j} &= D_{r, (\boldsymbol{\mu}_i, \boldsymbol{\mu}_j)} \qquad \forall r>1, \text{if } \boldsymbol{\mu}_i=\boldsymbol{\sigma}_r \text{ or }  \boldsymbol{\mu}_j=\boldsymbol{\sigma}_r \tag{6a} \label{eq5a}\\
    \hat{D}_{\boldsymbol{\mu}_i, \boldsymbol{\mu}_j} &= \eta_{\boldsymbol{\mu}_i, \boldsymbol{\mu}_j} + \beta \cdot \epsilon_{\boldsymbol{\mu}_i, \boldsymbol{\mu}_j} \qquad \forall (\boldsymbol{\mu}_i,\boldsymbol{\mu}_j) \in \Delta^{-} \times \Delta^{-}, \tag{6b} \label{eq5b}
\end{align*}
\normalsize

\noindent where the variables $\eta_{\boldsymbol{\mu}_i,\boldsymbol{\mu}_j}$ and $\epsilon_{\boldsymbol{\mu}_i,\boldsymbol{\mu}_j}$ are the mean and the standard deviation of the approximated distances between $\boldsymbol{\mu}_i$ and $\boldsymbol{\mu}_j$, obtained through the $R$ references. Eq.~(\ref{eq5a}) fills $\boldsymbol{\hat{D}}$ with the exact distances calculated in Eq.~(\ref{eq5pa}), and Eq.~(\ref{eq5b}) calculates the approximations when the exact distance is not available. $\beta$ is an input parameter that adjusts the underestimation or overestimation of the exact distances, and it is empirically adjusted by sampling exact distances within $\Delta^{-} \times \Delta^{-}$.

\subsection{Wasserstein-based kernels}
\label{sec:featurespace}

The success of kernel methods is driven by several factors: kernel functions facilitate the analysis of datasets in many applications, provide geometric interpretations, and produce computationally tractable models.  In this section, we give essential background on kernel functions, introduce kernels endowed with Wasserstein distances, and utilize kernel PCA for obtaining data-driven feature maps.

Formally, a kernel function $k(\cdot,\cdot)$ is any real-valued symmetric bivariate function that measures similarity within a non-empty dataset. A popular and versatile kernel function is the exponential kernel, which is used with diverse distance measures. Let us define $\mathcal{Y} := \{\boldsymbol{y}^1, \ldots, \boldsymbol{y}^N\}$, where $\boldsymbol{y}^j \in \mathbb{R}^d$ for every $j$. The exponential kernel is as follows:
\small
\begin{align*}
    k(\boldsymbol{y}^i, \boldsymbol{y}^j) = \exp{\left( - \gamma B_{\boldsymbol{y}^i, \boldsymbol{y}^j}^2 \right)} \qquad \forall (\boldsymbol{y}^i, \boldsymbol{y}^j) \in \mathcal{Y} \times \mathcal{Y}, \tag{7} \label{eq7}
\end{align*}
\normalsize

\noindent where $\gamma$ is an adjustable parameter, and $B_{\boldsymbol{y}^i, \boldsymbol{y}^j}$ is a distance measure between $\boldsymbol{y}^i$ and $\boldsymbol{y}^j$, e.g., $B_{\boldsymbol{y}^i, \boldsymbol{y}^j} = \|\boldsymbol{y}^i - \boldsymbol{y}^j \|_2$. 

A fundamental property of the kernel methods is defining the \textit{reproducing kernel Hilbert spaces} (RKHS). In brief, this property allows for mapping the input data $\mathcal{Y}$ to a high-dimensional \textit{feature space} when the kernel is p.d.:
\small
\begin{align*}
    k(\boldsymbol{y}^i, \boldsymbol{y}^j) &= \left \langle \phi(\boldsymbol{y}^i) , \phi(\boldsymbol{y}^j) \right \rangle \quad \forall (\boldsymbol{y}^i, \boldsymbol{y}^j) \in \mathcal{Y} \times \mathcal{Y}, \tag{8} \label{eq9k}
\end{align*}
\normalsize

\noindent where $\phi$ is a \textit{feature map} function that embeds the data into a dot product space. Furthermore, p.d. kernels have useful composition properties, where the addition or multiplication of p.d. kernels results in another p.d. kernel. This enables the simultaneous handling of different similarity metrics. For instance, let $k_1(\cdot,\cdot)$ and $k_2(\cdot,\cdot)$ be arbitraty p.d. kernels, and $\alpha_1, \alpha_2 \geq 0$ be arbitrary positive scalars, then the following kernels are p.d.:
\small
\begin{align*}
    &k_{p}(\boldsymbol{y}^i, \boldsymbol{y}^j) = k_{1}(\boldsymbol{y}^i, \boldsymbol{y}^j) \cdot  k_{2}(\boldsymbol{y}^i, \boldsymbol{y}^j) \tag{9a} \label{eqkmul}\\
    &k_{a}(\boldsymbol{y}^i, \boldsymbol{y}^j) = \alpha_1 k_{1}(\boldsymbol{y}^i, \boldsymbol{y}^j) +  \alpha_2 k_{2}(\boldsymbol{y}^i, \boldsymbol{y}^j), \tag{9b} \label{eqksum}
\end{align*}
\normalsize

\noindent in which $(\boldsymbol{y}^i, \boldsymbol{y}^j)$ is any given pair in $\mathcal{Y} \times \mathcal{Y}$. 

Positive definiteness holds in Eq.~(\ref{eq7}) for Euclidean distances, but it is not guaranteed for arbitrary distances
. In particular, if we have a set $\Delta = \{\boldsymbol{\mu}_l\}_{l=1}^S$ of discrete distributions, the Wasserstein-based exponential kernel defined below is known to be indefinite~\cite{de2020wasserstein}:
\small
\begin{align*}
    k(\boldsymbol{\mu}_i, \boldsymbol{\mu}_j) &= \exp{\left( - \gamma W(\boldsymbol{\mu}_i, \boldsymbol{\mu}_j)^2 \right)} \qquad \forall (\boldsymbol{\mu}_i, \boldsymbol{\mu}_j) \in \Delta \times \Delta. \tag{10} \label{eq10k} 
\end{align*}
\normalsize

In the following, we define a shifted Wasserstein-based kernel to handle the indefiniteness of Eq.~(\ref{eq10k}) and analyze the impact of the shift in clustering solutions.

\subsubsection{Constant shift Wasserstein-based kernel} \label{sec:pdkernelwass}

The following arguments apply to arbitrary kernels, though we focus on the kernel function in Eq.~(\ref{eq10k}). We define the corresponding kernel matrix on $\Delta$ as:
\small
\begin{align*}
    K_{ij} = k(\boldsymbol{\mu}_i,\boldsymbol{\mu}_j) \qquad \forall i=1,\ldots,S , j = 1,\ldots, S. \tag{11a} \label{eq10ak}
\end{align*}
\normalsize

A (strictly) p.d. kernel is a function that gives rise to a (strictly) p.d. kernel matrix. The matrix $\boldsymbol{K}$ is p.d. if it satisfies:
\small
\begin{align*}
    \sum_{i,j} c_i c_j K_{ij} \geq 0, \tag{11b} \label{eq10bk}
\end{align*}
\normalsize

\noindent for all $c_i \in \mathbb{R}$, where $c_i$ are arbitrary real scalars. If the equality in Eq.~(\ref{eq10bk}) only occurs for $\boldsymbol{c} = \boldsymbol{0}$, we call $\boldsymbol{K}$ strictly p.d. To avoid indefiniteness of $\boldsymbol{K}$, it is common practice to replace the kernel matrix by $\boldsymbol{K}' = \boldsymbol{K} + \varphi \boldsymbol{I}$, where $\varphi$ is a small positive constant called the \textit{jitter factor}
. Thus, the shifted kernel matrix $\boldsymbol{K}'$ is p.d. if the following inequality holds:
\small
\begin{align*}
    \sum_{i,j} c_i c_j K_{ij} + \sum_i c_i^2 \varphi \geq 0 \qquad \forall \boldsymbol{c} \in \mathbb{R}^S. \tag{11c} \label{eq10ck}
\end{align*}
\normalsize

It is usual to assign a small value $\varphi$. However, it is important to understand the consequences of shifting the kernel matrix. Here, we prove that kernel shifting results in invariant optimal assignments for the kernelized \texttt{k-medoids} problem with squared Euclidean distances. The \texttt{k-medoids} problem can be formulated using a kernel matrix:
\small
\begin{align*}
    \min \qquad &\sum_{i=1}^S \sum_{j=1}^S \left(K_{jj} + K_{ii} - 2K_{ij}\right) \vartheta_{ij} \tag{12a} \label{eq11acost}\\
    \text{s.t.} \qquad &\sum_{j=1}^S \rho_{j} = T \quad \text{;} \quad \sum_{j=1}^S \vartheta_{ij} = 1  \qquad \forall i \tag{12b} \label{eq11bcost}\\
    &\vartheta_{ij} \leq \rho_{j} \quad \text{;} \quad \vartheta_{ij}, \rho_{j} \in \{0,1\} \qquad \forall i, j \tag{12c} \label{eq11dcost}
\end{align*}
\normalsize

\noindent where $\vartheta_{ij}$ indicates the assignment of the $i$-th data entry to the $j$-th medoid, and the variable $\rho_{j}$ determines if the $j$-th data entry is a medoid. The number of data entries is $S$, the number of clusters is $T$, and $K_{ij}$ represents the similarity between $i$ and $j$. When the kernel matrix in Eq.~(\ref{eq11acost}) is replaced by a shifted kernel matrix, the objective function becomes:
\small
\begin{align*}
    2 \varphi (S-T) + \sum_{i=1}^S \sum_{j=1}^S \left(K_{ii} + K_{jj} - 2K_{ij}\right) \vartheta_{ij}. \tag{13} \label{eq12cost}
\end{align*}
\normalsize

The detailed steps for obtaining Eq.~(\ref{eq12cost}) are provided in the Supplementary Information, Section S1. We state the following theorem for the updated objective function.

\begin{theorem}
Given an arbitrary $S \times S$ kernel matrix, the optimal clustering assignments of the kernel \texttt{k-medoids} problem, Eqs.~(\ref{eq11acost})-(\ref{eq11dcost}), are invariant to a diagonal shift.
\end{theorem}

\begin{proof}
Since the cost function in Eq.~(\ref{eq12cost}) is equal to the cost function in Eq.~(\ref{eq11acost}) plus a constant, the minimization problems defined by these functions are equivalent. Therefore, the optimal clustering assignments remain the same.
\end{proof}

Let $\boldsymbol{D}$ be a matrix of pairwise Wasserstein distances for a set of distributions $\Delta$, which can be estimated as explained in Section~\ref{sec:refselection}. We define the shifted Wasserstein-based exponential kernel matrix as:
\small
\begin{align*}
    \boldsymbol{K}^W &= \exp \left(- \gamma \boldsymbol{D} ^{\circ 2}\right) + \varphi \boldsymbol{I} \tag{14} \label{eq12cker}
\end{align*}
\normalsize

\subsubsection{Kernel PCA} \label{sec:kernelpca}

A p.d. kernel can be expressed as the dot product of feature maps, as shown in Eq.~(\ref{eq9k}). This enables the generalization of PCA, one of the most widely used multivariate statistical techniques~\cite{ZHANG2024110591}, to the feature space, referred to as kernel PCA. Kernel PCA allows for obtaining a data-driven feature map from a dataset and selecting components based on explained variance.

Kernel PCA consists of two steps
. First, the kernel matrix $\boldsymbol{K}^{W} \in \mathbb{R}^{S \times S}$, defined over the dataset $\Delta$, must be centered as follows:
\small
\begin{align*}
    \tilde{K}_{\boldsymbol{\mu}_{i}, \boldsymbol{\mu}_{j}} = &K_{\boldsymbol{\mu}_{i}, \boldsymbol{\mu}_{j}}^{W} - \mathbb{E}_{\boldsymbol{\mu}}[K_{\boldsymbol{\mu}, \boldsymbol{\mu}_{j}}^{W} ] - \mathbb{E}_{\boldsymbol{\nu}}[K_{\boldsymbol{\mu}_{i},\boldsymbol{\nu}}^{W} ] + \mathbb{E}_{\boldsymbol{\mu}}[\mathbb{E}_{\boldsymbol{\nu}}[K_{\boldsymbol{\mu},\boldsymbol{\nu}}^{W} ]] \qquad \forall (\boldsymbol{\mu}_i, \boldsymbol{\mu}_j) \in \Delta \times \Delta. \tag{15} \label{eq11a} 
\end{align*}
\normalsize

Afterward, we obtain a feature map $\boldsymbol{\Phi} \in \mathbb{R}^{U \times S}$ by computing the eigendecomposition of $\tilde{\boldsymbol{K}}$. We use the eigenvectors $\boldsymbol{\upsilon}^1, \ldots, \boldsymbol{\upsilon}^S \in \mathbb{R}^{S}$ and eigenvalues $\lambda_1 \geq \ldots \geq \lambda_S$ to define the feature map:
\small
\begin{align*}
\boldsymbol{\Phi} = \boldsymbol{\Lambda}^{-1/2} \boldsymbol{V}^{\top} \tilde{\boldsymbol{K}}, \tag{16} \label{eq14u}
\end{align*}
\normalsize

\noindent where $\boldsymbol{\Lambda}^{-1/2} = \text{diag}([\lambda_1^{-1/2} \cdots \lambda_U^{-1/2} ])$, $\boldsymbol{V} = [\boldsymbol{\upsilon}^1 \cdots \boldsymbol{\upsilon}^U]$, and $U \leq S$ are the retained principal components. A popular criterion to select $U$ is the Kaiser rule
, which retains principal components with eigenvalues greater than 1 while discarding those associated with lower variance.


Unfortunately, the capabilities of kernel PCA come at a high computational cost. The time and memory complexities are $\mathcal{O}(S^3)$ and $\mathcal{O}(S^2)$, respectively. Hence, it is impractical for large-scale tasks, i.e., when $S$ reaches tens of thousands or higher values. A solution is to use the Nystr{\"o}m approximation for the eigendecomposition of $\tilde{\boldsymbol{K}}$. This method requires a subset of $M$ columns of $\tilde{\boldsymbol{K}}$~\cite{GUO2024110746} and is known to provide good-quality approximations even for $M$ on the order of hundreds~\cite{he2018kernel}. Additionally, it has time and memory complexities of $\mathcal{O}(S M^2 + M^3)$ and $\mathcal{O}(M^2)$, respectively.

Let $\tilde{\boldsymbol{K}}_{M \times M} \in \mathbb{R}^{M \times M}$ be a matrix with $M$ sampled columns and the corresponding rows from $\tilde{\boldsymbol{K}}$. The  Nystr{\"o}m method approximates the eigenvalues and eigenvectors of $\tilde{\boldsymbol{K}}$ as follows:
\small
\begin{align*}
\hat{\boldsymbol{\Lambda}} = \frac{S}{M} \boldsymbol{\Lambda}^{nys} \quad \text{;} \quad \hat{\boldsymbol{V}} = \sqrt{\frac{M}{S}} \tilde{\boldsymbol{K}}_{S \times M}  \boldsymbol{V}^{nys} \left(\boldsymbol{\Lambda}^{nys}\right)^{-1} \tag{17} \label{eq17bnys}
\end{align*}
\normalsize

\noindent where $\hat{\boldsymbol{\Lambda}}, \boldsymbol{\Lambda}^{nys}  \in \mathbb{R}^{M \times M}$ are the approximated eigenvalues matrices of $\tilde{\boldsymbol{K}}$ and $\tilde{\boldsymbol{K}}_{M \times M}$, respectively; similarly, $\hat{\boldsymbol{V}} \in \mathbb{R}^{S \times M}$ and $\boldsymbol{V}^{nys} \in \mathbb{R}^{M \times M}$ represent the corresponding eigenvectors matrices. $\tilde{\boldsymbol{K}}_{S \times M} \in \mathbb{R}^{S \times M}$ is the matrix with the $M$ sampled columns. In this way, the approximation of $\tilde{\boldsymbol{K}}$ is such that
$\tilde{\boldsymbol{K}} \approx  \hat{\boldsymbol{K}} =  \hat{\boldsymbol{V}} \hat{\boldsymbol{\Lambda}} \hat{\boldsymbol{V}}^{\top}$.

Nevertheless, the approximate eigenvectors $\hat{\boldsymbol{V}}$ are not orthogonal, so they do not define a proper PCA. Thus, we perform PCA on the approximate feature maps $\hat{\boldsymbol{\Phi}} \in \mathbb{R}^{M \times S}$ (computed with Eq.~(\ref{eq14u})) to obtain $M$ orthogonal directions. Moreover, $U \leq M$ components can be retained based on variance criteria, such as the Kaiser rule. The feature maps are then be retrieved using Eq.~(\ref{eq14u}).

\subsection{Clustering mapped features}
\label{sec:clustering}

After computing pairwise Wasserstein distances (Section \ref{sec:wassdist}) and deriving data-driven feature maps (Section \ref{sec:featurespace}), we can cluster the feature maps. We apply \texttt{k-medoids} to cluster the mapped features $\boldsymbol{\Phi} = [\boldsymbol{\xi}^1 \cdots \boldsymbol{\xi}^S] \in \mathbb{R}^{U \times S}$, where $\boldsymbol{\xi}^j \in \mathbb{R}^{U}$  for $j=1,\ldots,S$. This method allows for the identification of cluster medoids in the dataset. Furthermore, because the map $\boldsymbol{\Phi}$ consists of vector data, \texttt{k-medoids} can be applied directly, leading to straightforward computations.

However, the mapped features depend on the kernel matrix (Eq.~(\ref{eq14u})), which is derived from the kernel function used. This kernel may follow the form of Eq.~(\ref{eq7}), which requires selecting the parameter $\gamma$, or it may be composed of kernels as in Eqs.~(\ref{eqkmul})–(\ref{eqksum}), each requiring its own parameter. As kernel parameters affect the feature maps, selecting them carefully is crucial to ensure clustering robustness and quality.  To this end, we use validity indices to evaluate clustering results and guide the selection of kernel parameters. In Section~\ref{sec:resrobustness} and Section~\ref{sec:resshape}, we adopt a validity index to assess clustering robustness and develop another to evaluate cluster shape quality, respectively. In Section~\ref{sec:optimizingvalidity}, we show how to optimize kernel parameters based on these validity indices.

\subsubsection{Robustness assessment} \label{sec:resrobustness}

We evaluate the robustness of the clustering results across different initializations of \texttt{k-medoids}. High-quality results exhibit minimal variation in cluster assignments across initializations~\cite{vinh2010information}. To quantify this, we use the consensus index (CI), namely $I^{CI}$, based on the adjusted mutual information (AMI)
, which measures the similarity between different clustering solutions. The $I^{CI}$ has a stochastic range within the interval $[0, 1]$, with larger values indicating greater consensus among the solutions.

\subsubsection{Shape of clusters} \label{sec:resshape}

We develop a validity index based on sampling the Goodman-Kruskal (GK) index on subsets of the analyzed data to assess the shape quality of the clusters. This index is built on the concepts of \textit{concordant} and \textit{discordant} pairs of distances. A concordant distance pair is such that $||\boldsymbol{\xi}^{i_1} -\boldsymbol{\xi}^{i_2} ||_2 < ||\boldsymbol{\xi}^{i_3} -\boldsymbol{\xi}^{i_4} ||_2$, where $(\boldsymbol{\xi}^{i_1}, \boldsymbol{\xi}^{i_2})$ are in the same cluster and $(\boldsymbol{\xi}^{i_3}, \boldsymbol{\xi}^{i_4})$ are in different clusters. A discordant distance pair is one that satisfies $||\boldsymbol{\xi}^{i_1} -\boldsymbol{\xi}^{i_2} ||_2 > ||\boldsymbol{\xi}^{i_3} -\boldsymbol{\xi}^{i_4} ||_2$. A good clustering result is one where the number of concordant distance pairs dominates. The GK index is computed as: 
\small
\begin{align*}
    I^{GK} &= \frac{P^{c} - P^{d}}{P^{c} + P^{d}}, \tag{18} \label{eq15}
\end{align*}
\normalsize

\noindent with $P^c$ and $P^d$ being the count of all the concordant and discordant distance pairs, respectively. The $I^{GK}$ is bounded within the interval $[-1, 1]$, where higher values indicate better clustering. This index has the following characteristics: 1) it is robust against outliers because it is based on distance pairs
; 2) it performs well in high-dimensional settings~\cite{tomavsev2016clustering}
; and 3) it is distance-agnostic (as it only depends indirectly on distances). Nevertheless, the GK index computation becomes prohibitive for large data sets, as it requires comparing all possible distance pairs. We overcome this limitation by extending the GK index with a sampling procedure, as described in Algorithm~\ref{algorithm:fastgoodmankruskal}, which calculates our Fast Goodman-Kruskal index. A high-level description is given below. 

After initialization (Lines 1-6), the algorithm performs five steps. First, it samples $C$ point pairs $(\boldsymbol{\xi}^{i_1}, \boldsymbol{\xi}^{i_2})$ within the same clusters and stores them in a set $\Gamma^{a}$ (Lines 8-9). Second, it samples $C$ point pairs $(\boldsymbol{\xi}^{i_3}, \boldsymbol{\xi}^{i_4})$ from different clusters and stores them in a set $\Gamma^{b}$ (Lines 10-11)\footnote{Note that if there are repeated pairs, we continue sampling until there are $C$ unique pairs. For simplicity, this procedure is omitted in Algorithm~\ref{algorithm:fastgoodmankruskal}.}. Third, all possible distance pairs are obtained by combining the sampled pairs, i.e., $\Gamma^{a} \times \Gamma^{b}$, and placed in a set $\Gamma^{q}$ (Line 12). Fourth, the concordant and discordant distance pairs are counted (Lines 13-14). Fifth, the GK index is computed with Eq.~(\ref{eq15}) (Line 15). These steps are repeated $E$ times, and the sampled GK indices are averaged to obtain the FGK index (Line 18). The FGK index can be efficiently computed since its complexity depends on the number of sampled distance pairs and repetitions rather than the dataset size. 
\begin{algorithm}[ht]
\footnotesize
\caption{Fast Goodman-Kruskal index (FGK)}
\label{algorithm:fastgoodmankruskal}
\begin{algorithmic}[1]
\STATE Input: (1) The data set $\boldsymbol{\Phi} = [\boldsymbol{\xi}^1 \cdots \boldsymbol{\xi}^S$], (2) The vector with the cluster assignments $\boldsymbol{h}$, (3) The number of data pairs to be sampled $C$, and (4) The number of initializations for the sampling $E$.
\STATE Initialize: (1) The estimated validity index list $\hat{I}^{GK} \leftarrow \{\}$, and (2) The counter of the iterations $j \leftarrow 0$.
\STATE $\Theta \leftarrow$ Retrieve the unique cluster labels in $\boldsymbol{h}$ 

Obtain the labels of clusters with two or more points 
\STATE $\Xi \leftarrow \left\{t \in \Theta: \sum_{i=1}^S \mathbb{I}(\boldsymbol{h}_i = t) \geq 2 \right\}$ 

Compute the sampling weights of the clusters in $\Theta$
\STATE $\boldsymbol{w}^{a} \leftarrow \left[ \frac{1}{|\Theta|} \sum_{i=1}^S \mathbb{I}(\boldsymbol{h}_i = t) \right]_{t \in \Theta}$ 

Compute the sampling weights of the clusters in $\Xi$

\STATE $\boldsymbol{w}^{b} \leftarrow \left[ \frac{1}{|\Xi|} \sum_{i=1}^S \mathbb{I}(\boldsymbol{h}_i = t) \right]_{t \in \Xi}$  

\REPEAT
    \STATE $\mathcal{L} 
    \leftarrow$ Sample $C$ cluster label pairs from $\Theta$ without replacement using the sampling weights $\boldsymbol{w}^{a}$
    \STATE $\Gamma^{a} \leftarrow$ A pair of point entries from $\boldsymbol{\Phi}$ is sampled for each cluster label pair in $\mathcal{L}$
    \STATE $\boldsymbol{l}^{b} 
    \leftarrow$ Sample $C$ cluster labels from $\Xi$ with the weights $\boldsymbol{w}^{b}$
    \STATE $\Gamma^{b} \leftarrow$ Sample data pair entries from $\boldsymbol{\Phi}$ without replacement for each label in $\boldsymbol{l}^{b}$

    Compute all the distance pairs of data entries with a Cartesian product
    \STATE $\Gamma^{q} \leftarrow \Gamma^a \times \Gamma^b$
    
    Evaluate the number of concordant distance pairs
    \STATE  $P^c \leftarrow \quad \sum_{(i_1,i_2,i_3,i_4) \in \Gamma^q} \mathbb{I}\left(||\boldsymbol{\xi}^{i_1} -\boldsymbol{\xi}^{i_2} ||_2 < ||\boldsymbol{\xi}^{i_3} -\boldsymbol{\xi}^{i_4} ||_2 \right)$

    Evaluate the number of discordant distance pairs
    \STATE  $P^d \leftarrow \quad \sum_{(i_1,i_2,i_3,i_4) \in \Gamma^q} \mathbb{I}\left( ||\boldsymbol{\xi}^{i_1} -\boldsymbol{\xi}^{i_2} ||_2 > ||\boldsymbol{\xi}^{i_3} -\boldsymbol{\xi}^{i_4} ||_2 \right)$

    Calculate the sampled GK index
    \STATE  $\hat{I}^{GK} \leftarrow $ Append the result $\left( \frac{P^c - P^d}{P^c + P^d} \right)$
    \STATE $j \leftarrow j + 1$
\UNTIL{$j = E$}

Compute the Fast Goodman-Kruskal index
\STATE $I^{FGK} = \frac{1}{E} \sum_{j=1}^E \hat{I}^{GK}_j$
\RETURN $I^{FGK}$
\end{algorithmic} 
\end{algorithm}

\subsubsection{Optimizing the kernel parameters} \label{sec:optimizingvalidity}

Assume the kernel function used to obtain the kernel matrix and the feature maps in Eq.~(\ref{eq14u}) is composed of $F$ kernel functions, as defined in Eq.~(\ref{eq7}) (these can be composed as explained in Section~\ref{sec:featurespace}). The kernel parameters to be selected are $\boldsymbol{\gamma} = [\gamma^1, \cdots, \gamma^F]^{\top} \in \mathbb{R}^F$. To efficiently search for the kernel parameters $\gamma^i$ that result in high validity indices, lower and upper bounds, $\underline{\gamma^i}$ and $\overline{\gamma^i}$, must be set for each parameter. This is done by searching around the value $\gamma^{i}_{\max}$, which maximizes the off-diagonal variance of the $i$-th kernel matrix, as this value is expected to yield an informative kernel~\cite{merrill2020unsupervised}. The bounds can be set, for instance, as $\underline{\gamma^i} = 10^{-1/2} \cdot \gamma^{i}_{\max}$ and $\overline{\gamma^i} = 10^{1/2} \cdot \gamma^{i}_{\max}$. 

The kernel parameters $\boldsymbol{\gamma}$ are optimized to yield the best validity indices for a given number of clusters:
\small
\begin{align*}
    \max_{\underline{\boldsymbol{\gamma}} \leq \boldsymbol{\gamma} \leq \overline{\boldsymbol{\gamma}}} \quad \min_{I^{CI},I^{FGK}} \left\{I^{CI}, (I^{FGK} + 1)/2 \right\}, \tag{19} \label{maxbayes}
\end{align*}
\normalsize

\noindent where the objective is to maximize the minimum between the two validity indices after rescaling the FGK index to the [0,1] range, consistent with the CI index. This objective is set to simultaneously identify clusters that are both robust and well-shaped. Classical global optimization techniques cannot handle the optimization problem in Eq.~(\ref{maxbayes}). To address this, we employ a Bayesian optimization search~\cite{bayesoptpython} to find the maximizing parameters, guided by random samples within $\underline{\boldsymbol{\gamma}} \leq \boldsymbol{\gamma} \leq \overline{\boldsymbol{\gamma}}$.


\section{Experimental results}
\label{sec:experiments}

We demonstrate the effectiveness of our framework by clustering real world time series data and synthetic power distribution graphs (PDGs). Two labeled public time series datasets are analyzed: the twelve monthly power demand of Italy and pedestrian counts from Melbourne, available at the UCR archive~\cite{UCRtimearchive}. These datasets are used to evaluate the capability of the framework to identify known clusters correctly and to benchmark it against existing methods. In addition, the PDG datasets are nationwide and unlabeled models for Switzerland that consist of medium-voltage (MV) and low-voltage (LV) graphs~\cite{Oneto2024}. The aim of clustering the unlabeled PDGs is to select representative graphs that can reduce the complexity of the datasets by summarizing their cluster information, which is useful for power system applications. We implement all the methods in Python 3.10 and execute the experiments on a workstation with an AMD Ryzen Threadripper 3960X (24 cores) with 128 GB of RAM. The code is available at \url{https://github.com/aeonetos/wasserstein-kernel-clustering}.

\subsection{Time series}
\label{sec:timeseries}

Table~\ref{tab:datatime} shows the number of time series for each dataset, the number of labeled classes, and the duration and resolution of the samples. The Italy dataset consists of electrical power demand time series, and the classes distinguish days from October to March and April to September. The Melbourne dataset consists of pedestrian counts for 12 months, and the classes correspond to the location of the sensor placement.

\begin{table}[ht]
\footnotesize
\caption{Time series datasets. Duration and resolution are given in hours.}
\label{tab:datatime}
\centering
\captionsetup{justification=centering}
\begin{tabular}{l c c c c}
\toprule
 Name & Time series & Classes & Duration  & Resolution \\ 
 \hline Italy  & 1,096 & 2 & 24 & 1\\
 Melbourne  & 3,633 & 10 & 24 & 1\\
\bottomrule
\end{tabular}
\end{table}

\subsubsection{Wasserstein distance between Fourier power spectra} \label{sec:wassersteinfourier}

The time series analysis in the spectral domain is a common approach in which the time domain signal is represented by the energy distribution across frequencies in the spectral domain. The work in~\cite{Cazelles2021} introduces the Wasserstein distance between normalized power spectral densities (NPSDs) and its use in interpolation, augmentation, dimensionality reduction, and classification. Similarly, we use Wasserstein distances between NPSDs for time series clustering. We derive the NPSDs from normalized periodograms and obtain pairwise distances from the closed-form solution for one-dimensional densities. This process requires negligible computational effort. The distance calculations are performed on datasets normalized to the [0,1] range and on smoothed series, retaining 85\% of the variance using PCA.

\subsubsection{Kernel parametrization}
\label{sec:timekernels}

Exploratory data analysis shows that Italy's time series are bounded within similar ranges and have comparable average values, while the Melbourne dataset shows greater diversity. Consequently, we adopt two different kernels for their analysis. For the Italy dataset, the Wasserstein distances between the NPSDs provide enough information for clustering, as the problem consists mainly of shape detection. The corresponding kernel is defined as follows:
\small
\begin{align*}
    k_J(\boldsymbol{\tau}_i, \boldsymbol{\tau}_j) = \exp{\left( - \gamma^{J} W(\boldsymbol{\tau}_i, \boldsymbol{\tau}_j)^2 \right)}, \tag{20} \label{itakernel}
\end{align*}
\normalsize

\noindent where $(\boldsymbol{\tau}_i, \boldsymbol{\tau}_j)$ are the NPSDs of the signals $i$ and $j$, and $\gamma^{J}$ is the kernel parameter. 

To account for the amplitude variations in the Melbourne pedestrian counts, the corresponding Wasserstein kernel is composed of two additional kernels, which capture the instantaneous and total differences in pedestrian counts:
\small
\begin{align*}
    k_{M}(\mathcal{S}_i, \mathcal{S}_j) &= k_{J}(\mathcal{S}_i, \mathcal{S}_j) \cdot \left(k_T(\mathcal{S}_i, \mathcal{S}_j) + k_{A}(\mathcal{S}_i, \mathcal{S}_j)\right). \tag{21a} \label{melker}
\end{align*}
\normalsize

The arguments $(\mathcal{S}_i, \mathcal{S}_j)$ indicate the signal content from both the time and spectral domains, and each kernel extracts the corresponding information. The kernel $k_M$ is the pointwise multiplication of the shifted Wasserstein kernel with the sum of $k_{T}$ and $k_{A}$. The kernel $k_{T}$ captures the instantaneous similarity of the signals in the time domain, while $k_{A}$ accounts for the similarity in the total number of pedestrians over the signal duration. The rationale behind the composed kernel is supported by the idea that $k_J$ is adjusted according to the similarities in instantaneous and total pedestrian counts. The kernels $k_{T}$ and $k_{A}$ are defined as follows:
\small
\begin{align*}
    k_T(\mathcal{S}_i, \mathcal{S}_j) = \exp{\left( - \gamma^{T} ||\boldsymbol{x}_i - \boldsymbol{x}_j||_2^2 \right)}  \quad \text{;} \quad k_{A}(\mathcal{S}_i, \mathcal{S}_j) = \exp{\left( - \gamma^{A} |\boldsymbol{1}^{\top} (\boldsymbol{x}_i - \boldsymbol{x}_j)|^2 \right)}, \tag{21b} \label{addmelker}
\end{align*}
\normalsize

\noindent where $\boldsymbol{x}_i$ and $\boldsymbol{x}_j$ represent the vectors of the series $i$ and $j$ in the time domain, and $\boldsymbol{1}$ denotes the ones vector. The parameters $\gamma^T$ and $\gamma^A$ control the similarity measure for the corresponding kernels. A jitter factor of $10^{-3}$ is used in the kernels in Eq.~(\ref{itakernel}) and Eqs.~(\ref{melker})--(\ref{addmelker}) to prevent ill-conditioning. We use kernel PCA to mitigate the effect of low-variance features and apply the Kaiser rule to choose the number of components. Moreover, the eigendecomposition is solved exactly since the computations are tractable because the dataset sizes are on the order of $10^3$.

\subsubsection{Clustering benchmarking}
\label{sec:timeclustering}

We perform clustering with \texttt{k-medoids} with the alternate method on the mapped features using the scikit-learn-extra library~\cite{sklearnextra}. The kernel parameter for the Italy dataset is optimized by sampling 20 random parameters, followed by 20 iterations of Bayesian optimization. For the Melbourne dataset, the three kernel parameters are optimized using 50 random samples and 50 iterations of Bayesian optimization. For each iteration, clustering is conducted three times with different \texttt{k-medoids++} initializations. The FGK index is computed by sampling 35 independent sets of 100 pairs of points for each initialization using Algorithm~\ref{algorithm:fastgoodmankruskal}. Moreover, since the Italy dataset has two classes, the objective function for kernel optimization includes the effective number of constituents. This is done to avoid the trivial solution of grouping all samples in a single cluster. Finally, the search range for the parameters is set between  $\underline{\gamma^i} = 10^{-1} \cdot \gamma^{i}_{\max}$ and $\overline{\gamma^i} = 10^{1} \cdot \gamma^{i}_{\max}$, where $\gamma^{i}_{\max}$ is the maximizer of the off-diagonal variance of the kernel matrix.
\begin{table}[ht]
\footnotesize
\caption{Purity benchmarking for time series datasets (in \%).}
\label{tab:purity}
\centering
\captionsetup{justification=centering}
\begin{tabular}{l c c c c}
\toprule
 Name & GAN & GMM & HC & WK \\ 
 \hline Italy  & 76.60 $\pm$ 6.08 & 53.78 $\pm$ 0.00 & 51.43 & \textbf{77.66 $\pm$ 1.34}\\
 Melbourne  & 59.22 $\pm$ 3.03 & 58.17 $\pm$ 3.55 & 58.29 &  \textbf{66.59 $\pm$ 2.40} \\
\bottomrule
\end{tabular}
\end{table}

For benchmarking, we compare the results with two non-kernel methods, namely Gaussian Mixture Models (GMM) with full covariance and Hierarchical Clustering (HC) with ward linkage, implemented in scikit-learn~\cite{scikit-learn}. We also include results from a deep learning method based on a generative adversarial network (GAN) clustering~\cite{kim2021learning}. The numerical performance of the clustering algorithms is evaluated with the purity metric for the given number of classes~\cite{SU2024kercorrelation}, using the ground truth labels. This external validity index is the ratio of correctly clustered samples to the total number of samples, ranging from 0 to 1, with one corresponding to maximum purity. The methods are executed on 70\% partitions of the datasets to ensure consistency with~\cite{kim2021learning}, and the results correspond to five different initializations (except HC, which is deterministic). Table~\ref{tab:purity} reports the average and standard deviation of purity for each method. We observe that WK (proposed method) outperforms the other methods in both datasets. In fact, it is significantly superior to GMM and HC and performs similarly to GAN on the Italy dataset. On the Melbourne dataset, WK surpasses all three methods.
\begin{table}[ht]
\footnotesize
\caption{Execution times per clustering iteration for the time series datasets (in seconds).}
\label{tab:clusteringexecutionseries}
\centering
\captionsetup{justification=centering}
\begin{tabular}{l c c c c}
\toprule
 & Min. & Std. & Avg. & Max.  \\ 
 \hline Italy   & 0.70   & 0.21  & 1.78 & 2.12 \\
 Melbourne    & 4.48  & 0.31 & 4.99 & 6.17 \\
\bottomrule
\end{tabular}
\end{table}

Table~\ref{tab:clusteringexecutionseries} shows the execution time statistics for the WK method for the datasets. The execution times account for the kernel PCA step and the clustering of the mapped features with \texttt{k-medoids}. These results show that the clustering framework is both efficient and stable in its execution. Although the HC method completes clustering in milliseconds, and GMM in at most a few seconds, both yield lower purity results. The clustering task does not require real-time solutions, so this speed gain does not represent a meaningful advantage. In addition, GAN requires 50,000 training iterations and achieves similar purity on the Italy dataset but lower purity on the Melbourne dataset. The significant difference in iteration complexity further highlights the computational advantage of WK. In Supplementary Information, Section S2, we also provide results on the impact of retained variance in clustering purity when using variance thresholds for component selection.

\begin{figure}[ht]
\centering
\captionsetup{justification=centering}
    \includegraphics[width=0.65\textwidth]{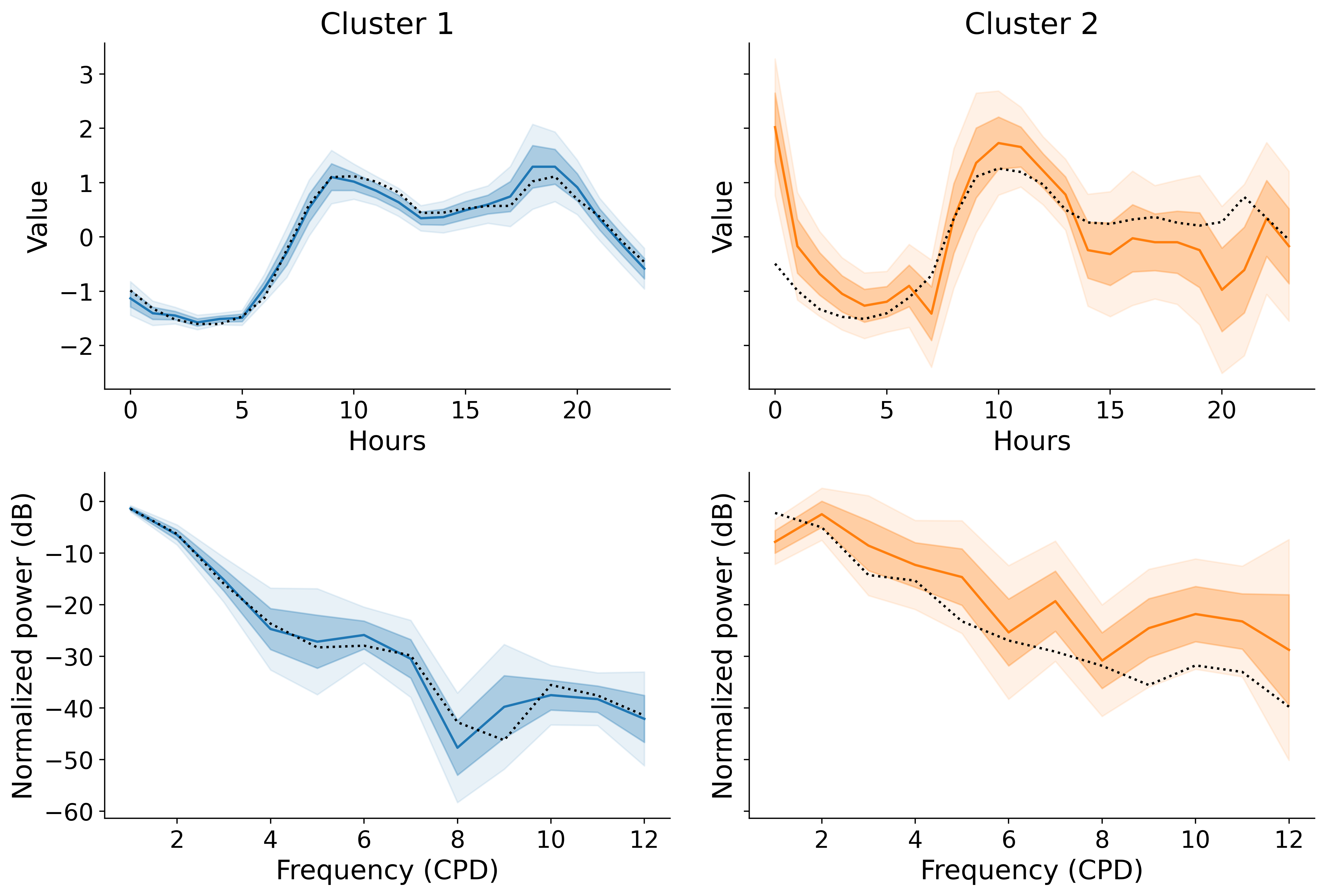}
    \caption{The clustering results for the Italy time series dataset. The top row shows the time domain, where we represent the median value for each cluster with dashed lines and display the medoids with solid lines. One and two standard deviations from the medoids are depicted with opaque bands. The time series are unitless and presented in their original, zero-centered format. The bottom row shows the frequency domain, including the medoids, the median values, and bands around the medoids of the NPSD at each frequency, expressed in CPD.}
\label{fig:italytime}
\end{figure}

In Figure~\ref{fig:italytime}, we present the clustering results for the Italy dataset, used to interpret the medoids and cluster structure in the time series. In the time domain, a clear pattern emerges: the first cluster exhibits significantly less variability than the second, as confirmed by the magnitude of the standard deviation at each time step. This small variability indicates that the first cluster is more homogeneous, and despite containing 39\% of the samples, a large portion of the dataset still resembles its medoid. We observe a similar pattern in the frequency domain, where the median normalized power of the first cluster closely matches that of its medoid. For the second cluster, variability is greater in both the time and frequency domains, which explains why the median lies at points outside one standard deviation from the medoid. Furthermore, analyzing the NPSD with the Wasserstein distance highlights two distinct cluster shapes. The frequency power in the first cluster decays rapidly, whereas the second cluster exhibits  periodicity up to four cycles per day (CPD), as indicated by the median values.

\subsection{Power distribution graphs}
\label{sec:powergraphs}


Table~\ref{tab:datagraph} shows the number of graphs for each dataset, their average number of nodes, and their standard deviation. To cluster the PDGs, we use information based on node properties (nodal voltage and power demand) and global properties (overall power demand and the number of nodes). Nodal voltages provide meaningful insights into the PDG state because they depend on AC power flows and line properties, indirectly conveying edge information, and nodal power demand provides consumption information. We represent graph nodes using discrete distributions of peak active power demand and voltage magnitude (p.u.), with all entries normalized between 0 and 1. Additionally, the global properties of each PDG have a vectorial representation.


\begin{table}[ht]
\footnotesize
\caption{Power distribution graph datasets.}
\label{tab:datagraph}
\centering
\captionsetup{justification=centering}
\begin{tabular}{l c c c}
\toprule
 Name & Graphs & Avg. nodes & Std. nodes \\ 
 \hline MV  & 879 & 113.38   & 49.72  \\
 LV   & 34,920 & 72.30  & 52.84\\
\bottomrule
\end{tabular}
\end{table}

\subsubsection{Wasserstein distance approximation}
 
We compute the Wasserstein distances between the discrete distributions using the method described in Section~\ref{sec:refselection}. We employ 25 reference distributions and calculate the exact distances with the Python POT library~\cite{flamary2021pot}. To tune the parameter $\beta$ in Eq.~(\ref{eq5b}), we sample 30,000 pairs of discrete distributions and compute the approximation error as:
\small
\begin{align}
    \left| \frac{\hat{D}_{\boldsymbol{\mu}_i, \boldsymbol{\mu}_j} - W(\boldsymbol{\mu}_i, \boldsymbol{\mu}_j)}{W(\boldsymbol{\mu}_i, \boldsymbol{\mu}_j)} \right|, \tag{22} \label{eq25error}
\end{align}
\normalsize

\noindent where $W(\boldsymbol{\mu}_i, \boldsymbol{\mu}_j)$ is the exact Wasserstein distance and $\hat{D}_{\boldsymbol{\mu}_i, \boldsymbol{\mu}_j}$ is the approximated distance. We test the values of $\beta$ in the range $[-1.5, 1.5]$ with a spacing of $0.5$, and select the one that minimizes the average error.

\begin{figure}[ht]
\centering
\captionsetup{justification=centering}
    \includegraphics[width=0.60\textwidth]{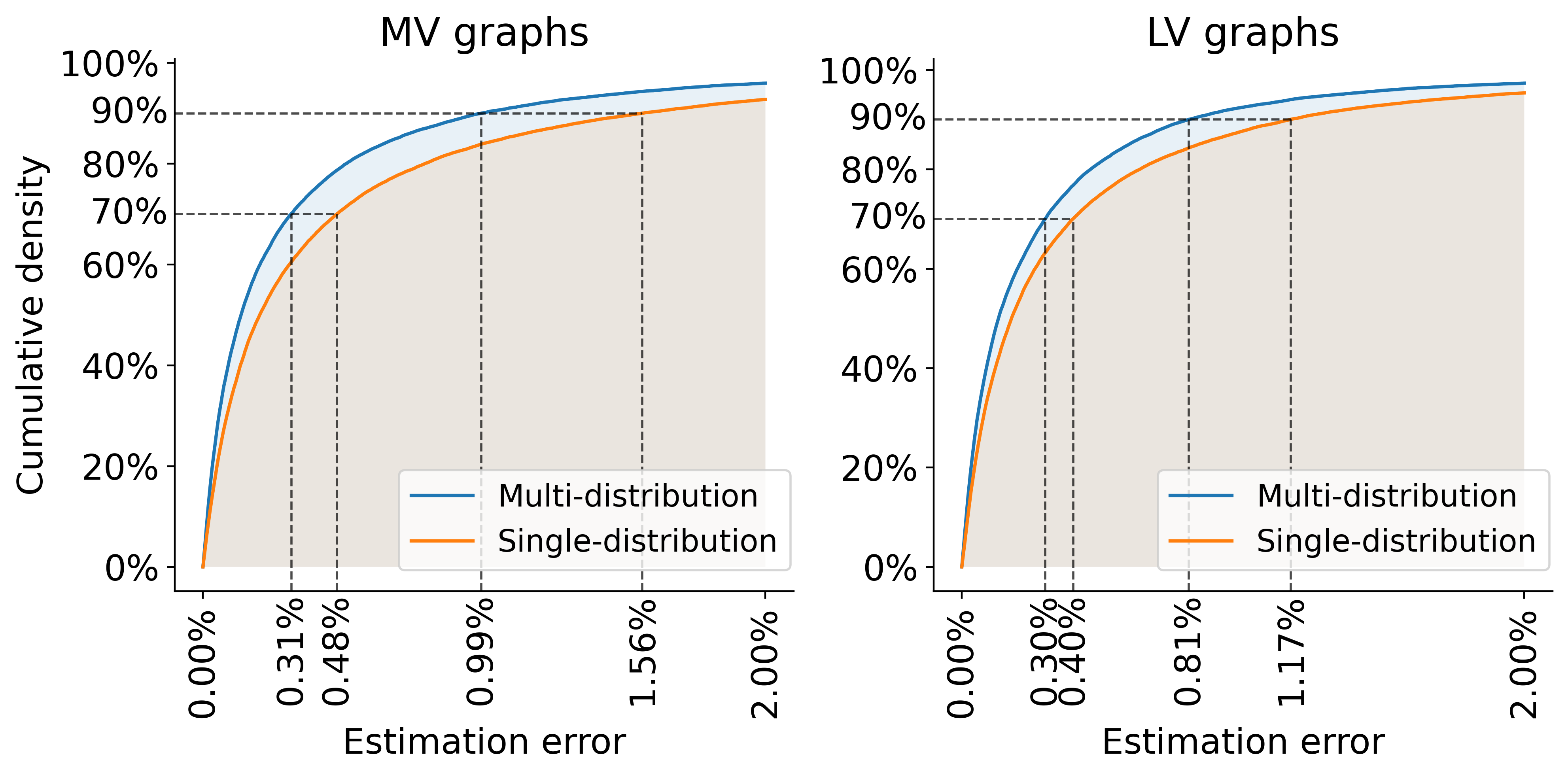}
    \caption{The cumulative density of approximation errors for the Wasserstein distance is shown for MV (left) and LV (right) graphs. Dashed lines indicate the 70th and 90th percentiles of the errors within each cumulative distribution.}
\label{fig:wassquality}
\end{figure}

We report the approximation errors for the method's initial and multiple-distribution approximations, i.e., \textit{Step 1} and \textit{Step 2} results, referred to as the single-distribution and multi-distribution approximations. The results for the multi-distribution approximation are reported for the best-approximating $\beta$ value, which is $-0.5$ for both the MV and LV datasets. Fig.~\ref{fig:wassquality} shows the cumulative density of the errors. The multi-distribution approximation outperforms the single-distribution approximation for both LV and MV graphs, with the difference becoming more pronounced for larger estimation errors. Furthermore, the multi-distribution approximation yields average errors of 0.435\% for the MV dataset and 0.380\% for the LV dataset, while the single-distribution approximation results in average errors of 0.640\% and 0.489\%, respectively. 
\begin{table}[ht]
\footnotesize
\caption{Execution times for complete pairwise Wasserstein distance calculations in the PDG datasets (in seconds).}
\label{tab:wassersteintime}
\centering
\captionsetup{justification=centering}
\begin{tabular}{l c c}
\toprule
 & Single distribution & Additional distributions \\ 
 \hline MV  & 2.34   & 30.71  \\
 LV    & 122.12  & 1,417.13\\
\bottomrule
\end{tabular}
\end{table}


Table~\ref{tab:wassersteintime} reports the wall-clock times for the corresponding calculations of the single distribution and additional distributions (computed in parallel). The additional distributions contain varying numbers of support vectors, with distributions having more support vectors resulting in longer execution times. Consequently, the time required to compute the single and additional distributions differs. Nevertheless, the total time needed for the multi-distribution approximation remains small: 33 seconds for the MV dataset and 1,539 seconds for the LV dataset (the sum of each row in Table~\ref{tab:wassersteintime}). These results demonstrate that the single-distribution's approximation quality can be further refined with additional distributions while maintaining a low computational burden.

\subsubsection{Kernel parametrization}\label{sec:kerparametrization}

The overall power demand of the PDGs and the number of nodes are essential characteristics, as they provide information about the power requirements imposed on substations and the number of consumers they serve. The Wasserstein distance between the discrete distributions does not account for these characteristics. Therefore, we compose the shifted Wasserstein-based kernel with a power demand kernel and a node number kernel as follows:
\small
\begin{align*}
    k_{G}(\mathcal{G}_i, \mathcal{G}_j) &= k_W(\mathcal{G}_i, \mathcal{G}_j) \cdot \left(k_P(\mathcal{G}_i, \mathcal{G}_j) + k_{V}(\mathcal{G}_i, \mathcal{G}_j)\right). \tag{23a} \label{eq19a}
\end{align*}
\normalsize

The kernel function $k_{G}$ between two PDGs is the pointwise multiplication of the shifted Wasserstein-based kernel, $k_W$, with the sum of the other two kernels, $k_P$ and $k_{V}$. The kernel $k_P$ captures the similarity in their overall power requirements, while $k_{V}$ captures the similarity in the number of nodes. The kernels' arguments are indicated as $(\mathcal{G}_i, \mathcal{G}_j)$, and each kernel extracts the corresponding information from the graphs. The composition of these kernels is based on the rationale that $k_W$ is adjusted based on the similarities in total power demand and the number of nodes. The expressions for $k_P$ and $k_{V}$ are introduced below:
\small
\begin{align*}
    k_P(\mathcal{G}_i, \mathcal{G}_j) = \exp{\left( - \gamma^{P} |\theta^{P}_{\mathcal{G}_i}-\theta^{P}_{\mathcal{G}_j}|^2 \right)}  \quad \text{;} \quad k_{V}(\mathcal{G}_i, \mathcal{G}_j) = \exp{\left( - \gamma^{V} |\theta^{V}_{\mathcal{G}_i}-\theta^{V}_{\mathcal{G}_j}|^2 \right)}, \tag{23b} \label{eq19c}
\end{align*}
\normalsize

\noindent where $\theta^{P}_{\mathcal{G}}$ and $\theta^{V}_{\mathcal{G}}$ represent the power demand and the number of nodes in the graph $\mathcal{G}$, and  $\gamma^{P}$ and $\gamma^{V}$ are the kernel parameters, respectively. To prevent ill-conditioning of the kernel in Eq.~(\ref{eq19a}), we shift $k_W$, $k_P$, and $k_{V}$ by a jitter factor of $10^{-3}$. We use kernel PCA to reduce the effect of low-variance features, retaining the number of components determined by the Kaiser rule. Additionally, we employ the Nystr\"om method for the LV dataset, sampling 5,000 columns to achieve a negligible reconstruction error.

\subsubsection{Clustering evaluation} \label{sec:kertuning}

The clustering is performed with \texttt{k-medoids} on the mapped features. We apply the PAM method for the smaller MV dataset and the alternate method for the LV dataset. The latter method offers significantly faster execution on large datasets. The kernel parameters are optimized by sampling 50 random combinations of parameters, followed by 50 iterations of Bayesian optimization. As with the time series datasets, we use three different \texttt{k-medoids++} initializations, and the FGK index is computed with 35 samplings of 100 pairs of points. The search ranges for the parameters are set to $\underline{\gamma^i} = 10^{-1/2} \cdot \gamma^{i}_{\max}$ and $\overline{\gamma^i} = 10^{1/2} \cdot \gamma^{i}_{\max}$. High-quality solutions may lie outside these ranges; thus, the approach is combined with grid search. We use context-dependent relative validity indices, as the FGK and CI validity indices are agnostic to contextual knowledge.

The context-dependent validity indices assess whether clustering based on the composed kernel simultaneously captures: 1) the number of nodes and total power demand, 2) the nodal voltage distributions, and 3) the normalized nodal power demand distribution\footnote{A graph's normalized nodal power demand distribution is computed as the percentage contribution of each node to the overall power demand.}. We first introduce the clustering approaches for building the validity indices, which consider each information type separately. The first method, Power and Nodes (PN), clusters the graphs according to the power demand and the number of nodes of the PDGs. We define the vectors $[\theta_{\mathcal{G}}^P, \theta_{\mathcal{G}}^{V}]^{\top}$ for every PDG, normalize them with min-max normalization to make the entries comparable, and cluster them with \texttt{k-medoids}. The second method, Wasserstein Voltage (WV), involves computing all the pairwise distances of the nodal voltage distributions using the closed-form univariate solution of the Wasserstein distance. These distances are then used to cluster the PDGs with \texttt{k-medoids}. The third approach, Wasserstein Power (WP), follows the same steps as the second one but uses the normalized nodal power demand distribution instead. 

We utilize a modified version of the commonly used Davies-Bouldin (DB) index, which measures the dispersion within and separation between clusters, with lower values being preferred. We compute the DB index under three separation measures: 1) $d_{PN}$, the Euclidean distance of the normalized vectors, 2) $d_{WV}$, the Wasserstein distance between nodal voltage distributions, and 3) $d_{WP}$, the Wasserstein distance between normalized nodal power demand distributions. We refer to the DB index based on these separation measures as $\varepsilon_{PN}$, $\varepsilon_{WV}$, and $\varepsilon_{WP}$, respectively. For each model presented, we compute the DB index for each separation measure. For instance, we compute $\varepsilon^{PN}_{WV}$, which is the DB index under the $d_{WV}$ separation measure for the PN clustering. We denote the minimum and maximum DB index attained by the clustering methods for a given separation measure, such as $d_{WV}$, as $\underline{\varepsilon_{WV}}$ and $\overline{\varepsilon_{WV}}$, respectively. Then, we define the normalized DB indices as:
\small
\begin{align}
    \psi_{PN} &= (\varepsilon_{PN} - \underline{\varepsilon_{PN}})/(\overline{\varepsilon_{PN}} - \underline{\varepsilon_{PN}}), \tag{24a} \label{eq23a} \\
    \psi_{WV} &= (\varepsilon_{WV} - \underline{\varepsilon_{WV}})/(\overline{\varepsilon_{WV}} - \underline{\varepsilon_{WV}}), \tag{24b} \label{eq23b} \\
    \psi_{WP} &= (\varepsilon_{WP} - \underline{\varepsilon_{WP}})/(\overline{\varepsilon_{WP}} - \underline{\varepsilon_{WP}}), \tag{24c} \label{eq23c}
\end{align}
\normalsize

\noindent where the variables $\psi_{PN}$, $\psi_{WV}$, and $\psi_{WP}$ are the normalized DB indices obtained under the separation measures $d_{PN}$, $d_{WV}$, and $d_{WP}$, respectively. Lastly, we select the WK clustering assignment with the lowest maximum and average normalized DB scores.

To find a small set of PDGs that represent the datasets and reduce dataset complexity, we evaluate 10 clusters, providing a small enough set of medoids while still conveying diverse information\footnote{In the evaluation process, we tested different numbers of clusters and could not observe a clear peak in the validity indices, even when testing up to 100 clusters. Having so many clusters would no longer effectively reduce dataset complexity.}. In the following, we assess the clustering execution performance, analyze the obtained validity indices, select clustering assignments, and measure their quality.
\begin{table}[ht]
\footnotesize
\caption{Execution times per clustering iteration for the power distribution graph datasets (in seconds).}
\label{tab:clusteringtime}
\centering
\captionsetup{justification=centering}
\begin{tabular}{l c c c c}
\toprule
 & Min. & Std. & Avg. & Max.  \\ 
 \hline MV   & 4.52   & 0.41  & 5.35 & 6.76    \\
 LV    & 242.25  & 5.80 & 253.08 & 274.49 \\
\bottomrule
\end{tabular}
\end{table}

%
%
The execution time statistics for the clustering processes are reported in Table~\ref{tab:clusteringtime}. These statistics correspond to the 50 kernel parameter initializations and the 50 Bayesian optimization iterations. The reported execution time includes the time taken to obtain the feature maps through kernel PCA and to perform \texttt{k-medoids} clustering on these maps. On average, the process takes 5.35 seconds for the MV graphs and 253.08 seconds for the LV graphs. Additionally, the standard deviations and the difference between the maximum and minimum times are minor. The clustering execution proves to be efficient, and the time scales linearly with respect to the number of graphs clustered. It is important to note that the execution time is independent of the graph sizes.
\begin{figure}[ht]
\centering
\captionsetup{justification=centering}
    \includegraphics[width=0.60\textwidth]{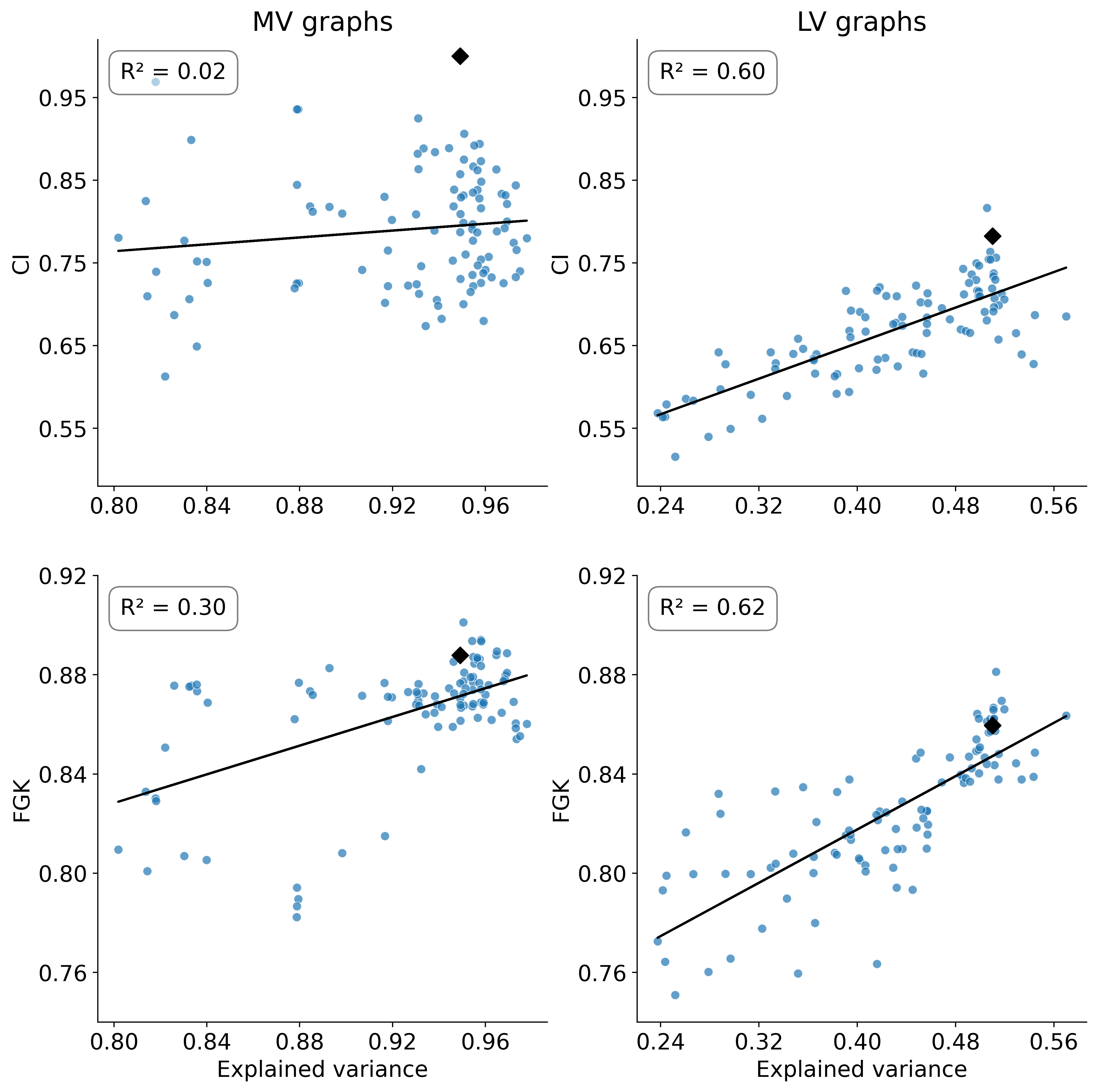}
    \caption{Validity indices plotted against the explained variance of the first five components. The validity indices corresponding to the selected clustering results are indicated with black diamonds.}
\label{fig:variancevsvalidity}
\end{figure}

Next, we analyze the relationship between the explained variance of the feature maps and the validity indices. Figure~\ref{fig:variancevsvalidity} reports the validity indices CI and FGK plotted against the explained variance of the first five components of the kernel PCA\footnote{Note that, for the LV graphs dataset, the retained variance corresponds to the Nystr{\"o}m approximation of the kernel matrix.}. For the MV graphs dataset, the CI shows no clear relationship with the explained variance. However, a subtle linear trend is observed between the FGK and the variance. On the other hand, the LV graphs dataset displays a linear relationship between CI/FGK values and the variance. Additionally, MV graphs generally exhibit higher CI and FGK values than the LV graphs. This difference can be attributed to the fact that the LV dataset requires more principal components than the MV dataset to explain the same variance. Furthermore, the difference in CI values is influenced by the \texttt{k-medoids} PAM method being more stable (used for the MV graphs) than the faster alternate method (used for the LV graphs).

\begin{table}[ht]
\footnotesize
\caption{Normalized DB indices of MV graphs clustering.}
\label{tab:comparisonMV}
\centering
\captionsetup{justification=centering}
\begin{tabular}{l c c c c}
\toprule
 & WK & PN & WP & WV \\ 
 \hline Voltage   & 0.227   & 0.796    & 1  & 0  \\
 Power    & 0.433  & 0.208     & 0      & 0.781  \\
Power\&Nodes   & 0.115  & 0     & 0.430      & 1  \\
\hline Average   & 0.258  & 0.335     & 0.477      & 0.594    \\
 Max    & 0.433   & 0.796  & 1      &  1\\
\bottomrule
\end{tabular}
\end{table}

\begin{table}[ht]
\footnotesize
\caption{Normalized DB indices of LV graphs clustering.}
\label{tab:comparisonLV}
\centering
\captionsetup{justification=centering}
\begin{tabular}{l c c c c}
\toprule
 & WK & PN & WP & WV \\ 
 \hline Voltage   & 0.232   & 0.826    & 1  & 0  \\
 Power    & 0.595  & 0.572     & 0      & 0.922  \\
Power\&Nodes   & 0  & 0     &  1      & 0.042  \\
\hline Average   & 0.277  & 0.466     & 0.667      & 0.321    \\
 Max    & 0.595   & 0.826  & 1      &  0.922 \\
\bottomrule
\end{tabular}
\end{table}

We sort the clustering solutions, corresponding to the 50 kernel parameter initializations and the 50 Bayesian optimization iterations. From the sorted solutions, we compute the context-dependent validity index in Eqs.~(\ref{eq23a})-(\ref{eq23c}) for the three solutions with the highest objective values. Afterward, the clustering solutions selected for the MV and LV cases are those that simultaneously yield the lowest average and maximum normalized DB scores. Table~\ref{tab:comparisonMV} and Table~\ref{tab:comparisonLV} report the normalized DB scores of the selected solutions for the MV and LV graphs, respectively.
The MV clustering has an FGK value of 0.888 and a CI of 1, while the LV clustering has an FGK of 0.860 and a CI of 0.782 (indicated by the black diamonds in Figure~\ref{fig:variancevsvalidity}). Furthermore, the WK method achieves the lowest average and maximum indices for both datasets compared to the other methods. Thus, the WK effectively incorporates the information of the nodal voltage distribution, the nodal power distribution, the graph's total power demand, and the number of nodes.
\begin{figure}[ht]
\centering
\captionsetup{justification=centering}
    \includegraphics[width=0.55\textwidth]{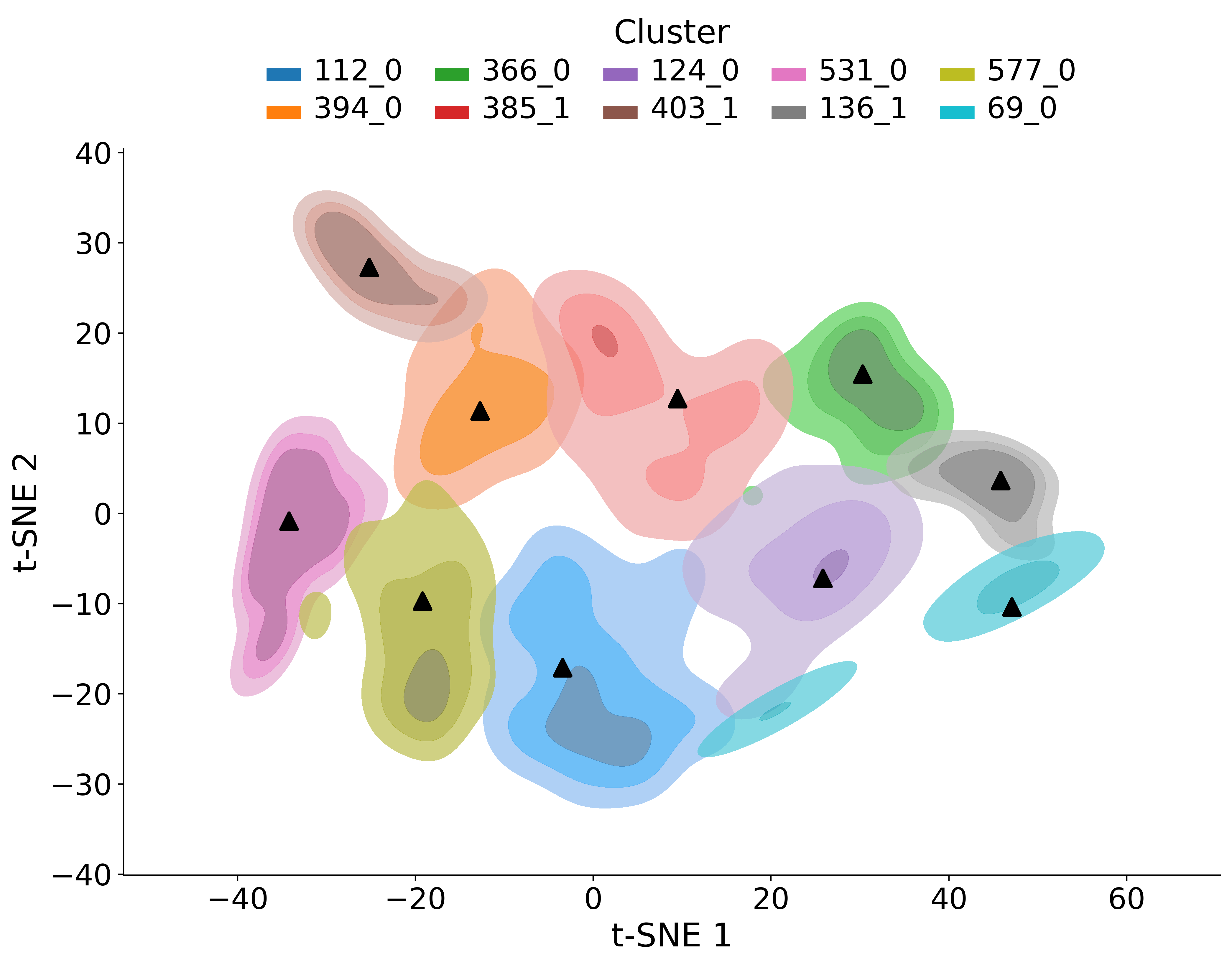}
    \caption{Projection via t-SNE of the MV graphs feature maps. The cluster medoids are indicated with a black triangle. The clusters are named after the identifier of the medoid in the dataset.}
\label{fig:tsneMV}
\end{figure} 

\begin{figure}[ht]
\centering
\captionsetup{justification=centering}
    \includegraphics[width=0.55\textwidth]{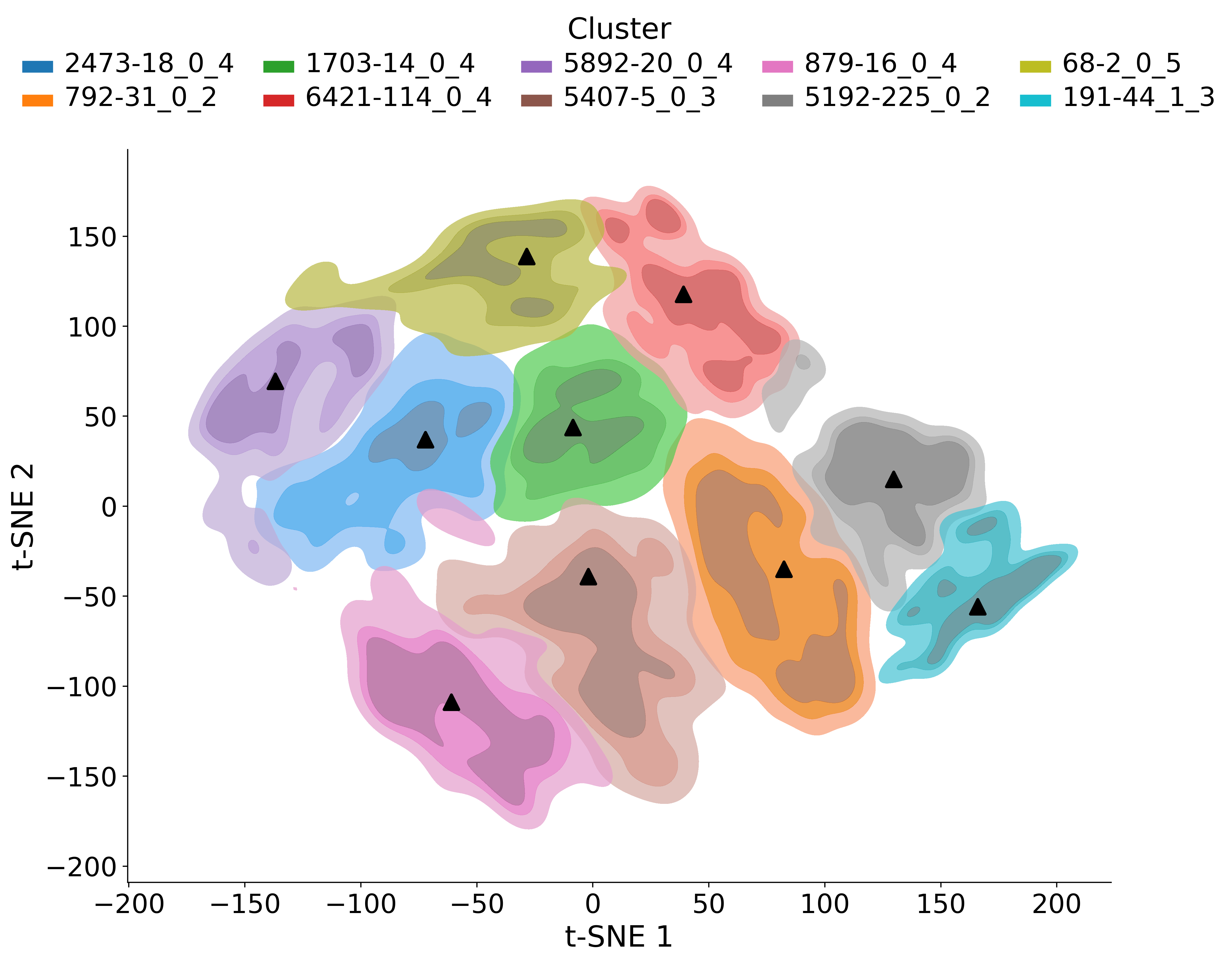}
    \caption{Projection via t-SNE of the LV graphs feature maps. The cluster medoids are indicated with a black triangle. The clusters are named after the identifier of the medoid in the dataset.}
\label{fig:tsneLV}
\end{figure} 
Figures~\ref{fig:tsneMV}~and~\ref{fig:tsneLV} display the t-SNE projection of the clusters for the MV and LV datasets, respectively. Specifically, the MV feature maps have 14 dimensions, while the LV feature maps have 222 dimensions, both of which are projected into two dimensions. Figures~\ref{fig:tsneMV}~and~\ref{fig:tsneLV} show density plots with contour levels, corresponding to kernel density estimations over the projected points, drawn with a threshold of 80\% of the probability mass. Three different color levels represent areas of equal probability mass for each cluster. Additionally, the medoids are indicated with black triangles. The clusters are labeled according to the name of their medoids in the corresponding datasets. 

We observe that the clusters show only slight overlaps and are well-distributed in the projected spaces. Although these visualizations are lower-dimensional projections, they confirm that the clusters are compact and exhibit clear separation. This outcome is expected given the high FGK values. Moreover, the medoids tend to lie close to or within the contour level of the highest density in both figures. This result indicates that the medoids display central characteristics of their respective clusters.
\begin{figure*}[t]
\centering
\captionsetup{justification=centering}
    \includegraphics[width=1.0\textwidth]{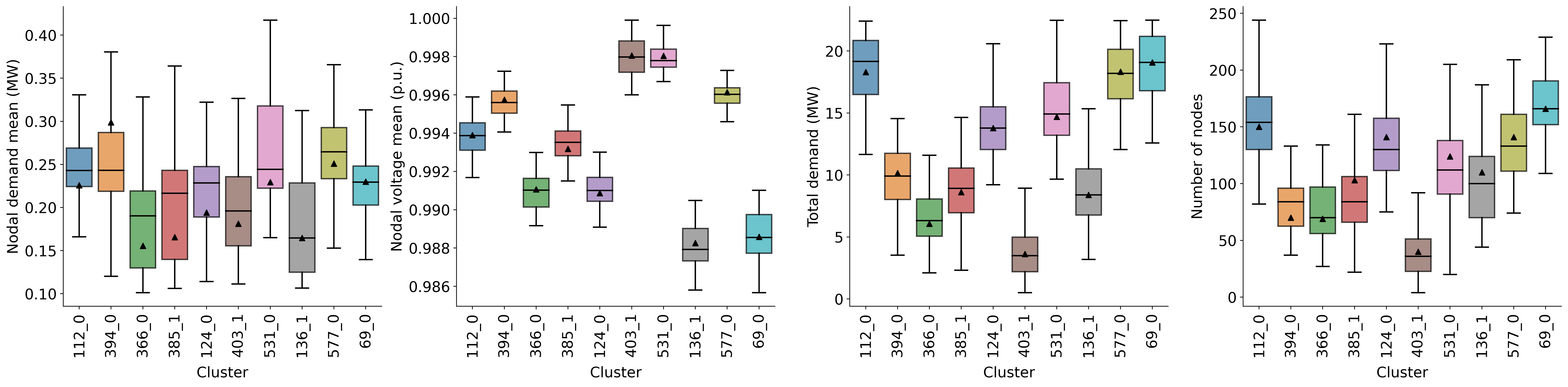}
    \caption{Distributions of MV graphs by cluster, including mean nodal demand, mean nodal voltage, total demand, and the number of nodes. The medoids of each cluster are marked with black triangles.}
\label{fig:mvboxstats}
\end{figure*}  

\begin{figure*}[t]
\centering
\captionsetup{justification=centering}
    \includegraphics[width=1.0\textwidth]{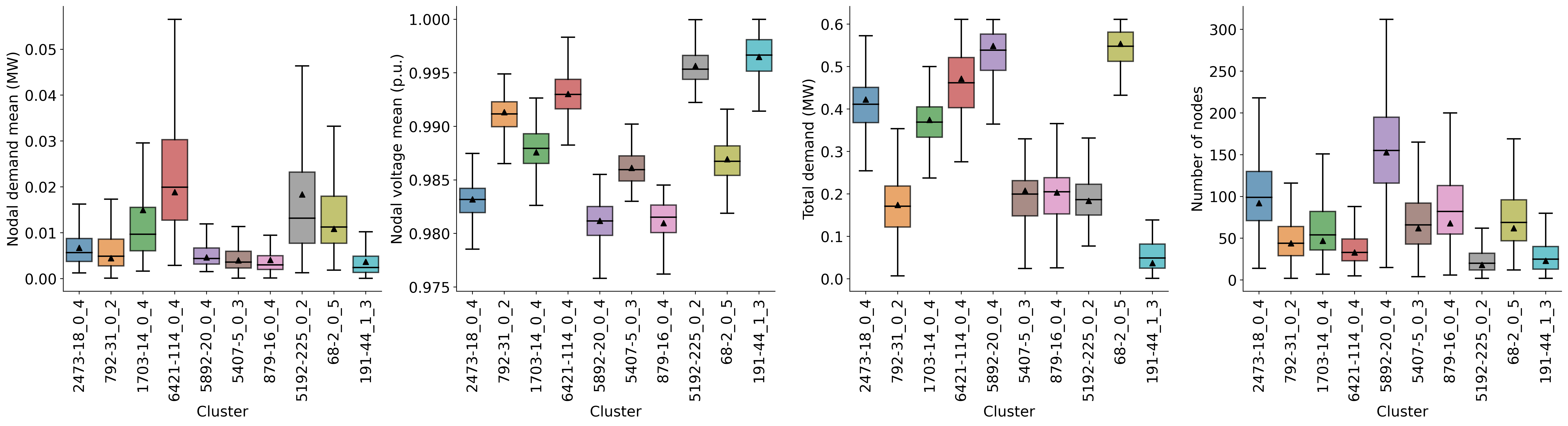}
    \caption{Distributions of LV graphs by cluster, including mean nodal demand, mean nodal voltage, total demand, and the number of nodes. The medoids of each cluster are marked with black triangles.}
\label{fig:lvboxstats}
\end{figure*}    

We further examine the clusters with respect to 1) the mean nodal demand, 2) the mean nodal voltage, 3) the total demand, and 4) the number of nodes. Figures~\ref{fig:mvboxstats}~and~\ref{fig:lvboxstats} show these properties for the MV and LV graphs, respectively, using the same colors as in Figures~\ref{fig:tsneMV}~and~\ref{fig:tsneLV}. We observe that most of the medoids, indicated by black triangles, lie within the interquartile ranges and close to the medians, which is expected if the medoids are representative objects of their clusters. Moreover, we found that the t-SNE axis 2 in Fig.~\ref{fig:tsneMV} is related to the total demand and number of nodes in the MV graphs. Higher values on this axis correspond to lower total demand and fewer nodes, as observed in Fig.~\ref{fig:mvboxstats}. Fig.~\ref{fig:tsneMV} shows that for the t-SNE axis 1, the higher values correspond to lower nodal voltage means, similarly as in Fig.~\ref{fig:mvboxstats}. For the LV graphs, we draw similar insights. The t-SNE axis 2 of Fig.~\ref{fig:tsneLV} is related to the total demand of the graphs since higher values on this axis correspond to higher total demand (shown in Fig.~\ref{fig:lvboxstats}). For the t-SNE axis 1, we observe that higher values are related to higher nodal voltage means. 


\section{Discussion and limitations}
\label{sec:discussion}

The results, along with their limitations, highlight avenues for future research. First, we use the Kaiser criterion to choose the number of components in kernel PCA. Nevertheless, it may be valuable to investigate alternative component selection methods and determine whether there is an optimal number of components to retain. 

Second, although we base the framework on geometrical notions and its operation can be explained, the clustering results could benefit from post-hoc interpretability methods. We analyzed the clustering assignments in the experiments to gain insights, but we could extend these interpretations by incorporating feature importance techniques.

Third, manifold learning techniques have provided valuable resources for improving clustering solutions in other works. A valuable extension of this work would be to incorporate these techniques into the feature maps derived from kernel PCA. Future developments could utilize them to enhance the framework’s efficacy.

Fourth, the PDG experiments considered distributional information over the nodes, specifically voltages and power demand. Practitioners can also use other unsupervised techniques to embed graph information into bag-of-vector representations, built on node embedding functions or Laplacian embeddings. These embeddings could provide further information to group graphs and identify prototypes.

Lastly, the discrete distributions in the PDG case study contain a number of support vectors on the order of $10^2$ (see Table~\ref{tab:datagraph}), for which computing the exact distances with the references is manageable. However, some applications may need an approximation of the discrete distributions, such as particle approximation.


\section{Conclusions}
\label{sec:conclusions}

This article proposes a clustering framework that integrates kernel methods with Wasserstein distances to analyze distributional data. This data type is prevalent in modern applications, including graph-structured data (e.g., electrical networks, brain functional networks, protein structures), power spectral densities of time series, and chemical compound representations. Therefore, the framework is valuable across domains, as demonstrated by the experimental results on time series and power distribution graphs. We have benchmarked the framework against existing models and showcased the summarizing capabilities of identified cluster prototypes. The execution performance also highlighted its computational tractability.

We have achieved fast and accurate estimations of pairwise Wasserstein distances in large, multi-dimensional datasets by developing an efficient approximation that leverages multiple reference distributions. Moreover, our introduction of shifted positive definite kernel functions based on Wasserstein distances enables well-conditioned compositions with other similarity measures and kernel PCA. In addition, we have leveraged scalable and distance-agnostic validity indices to optimize kernel parameters.

Nevertheless, our study has limitations that present promising avenues for future research. In particular, optimized component selection in kernel principal component analysis and manifold learning could improve the performance of the clustering. Embedding techniques for graphs and post-hoc analysis could improve the representational and interpretability capabilities of the framework.


\section*{Acknowledgments}
The research published in this publication was carried out with the support of the Swiss Federal Office of Energy SFOE as part of the SWEET EDGE project. 

The authors thank Enrico Ampellio for his valuable suggestions while preparing the manuscript. They also acknowledge the constructive comments from the anonymous reviewers, which have greatly improved the work.

\bibliographystyle{elsarticle-num-names}
\bibliography{refs}

\end{document}